\documentclass[12pt]{iopart}
\usepackage{graphicx,times,amssymb,cite,subfigure,stfloats,booktabs,multirow,bm}
\usepackage[lined,ruled,commentsnumbered]{algorithm2e}

\begin{document}

\title{Universal Adversarial Perturbations for CNN Classifiers in EEG-Based BCIs}
\author{Zihan~Liu$^{1,4}$, Lubin~Meng$^{1,4}$, Xiao~Zhang$^{1}$, Weili Fang$^2$ and Dongrui~Wu$^{1,3}$}
\address{$^1$ Key Laboratory of the Ministry of Education for Image Processing and Intelligent Control, School of Artificial Intelligence and Automation, Huazhong University of Science and Technology, Wuhan 430074, China.}
\address{$^2$ School of Design and Environment, National University of Singapore, 117566 Singapore.}
\address{$^3$ Zhejiang Lab, Hangzhou 311121, China.}
\address{$^4$ These authors contributed equally to this work.}
\ead{\mailto{zhliu95@hust.edu.cn}, \mailto{lubinmeng@hust.edu.cn}, \mailto{xiao\_zhang@hust.edu.cn},
	\mailto{bdgfw@nus.edu.sg},	\mailto{drwu@hust.edu.cn.}}
\maketitle

\begin{abstract}
Multiple convolutional neural network (CNN) classifiers have been proposed for electroencephalogram (EEG) based brain-computer interfaces (BCIs). However, CNN models have been found vulnerable to universal adversarial perturbations (UAPs), which are small and example-independent, yet powerful enough to degrade the performance of a CNN model, when added to a benign example. This paper proposes a novel total loss minimization (TLM) approach to generate UAPs for EEG-based BCIs. Experimental results demonstrated the effectiveness of TLM on three popular CNN classifiers for both target and non-target attacks. We also verified the transferability of UAPs in EEG-based BCI systems. To our knowledge, this is the first study on UAPs of CNN classifiers in EEG-based BCIs. UAPs are easy to construct, and can attack BCIs in real-time, exposing a potentially critical security concern of BCIs.
\end{abstract}

\vspace{2pc}

\section{Introduction} \label{sect:Intro}

A brain-computer interface (BCI) enables people to interact directly with a computer using brain signals. Due to its low-cost and convenience, electroencephalogram (EEG), which records the brain's electrical activities from the scalp, has become the most widely used input signal in BCIs. There are several popular paradigms in EEG-based BCIs, e.g., P300 evoked potentials~\cite{P300Sutton, P3001988,drwuTHMS2017,drwuTNSRE2016}, motor imagery (MI)~\cite{MI2001}, steady-state visual evoked potentials (SSVEP)~\cite{SSVEPSurvey}, etc.

Deep learning, which eliminates manual feature engineering, has become increasingly popular in decoding EEG signals in BCIs. Multiple convolutional neural network (CNN) classifiers have been proposed for EEG-based BCIs. Lawhern \emph{et al.}~\cite{EEGNet} proposed EEGNet, a compact CNN model demonstrating promising performance in several EEG-based BCI tasks. Schirrmeister \emph{et al.}~\cite{MNE} proposed a deep CNN model (DeepCNN) and a shallow CNN model (ShallowCNN) for EEG classification. There were also studies that converted EEG signals to spectrograms or topoplots and then fed them into deep learning classifiers~\cite{EEG2Image,DeepLearningMI,Tayeb2019}. This paper focuses on CNN classifiers which take raw EEG signals as the input, but our approach should also be extendable to other forms of inputs.

Albeit their promising performance, it was found that deep learning models are vulnerable to adversarial attacks~\cite{AdvExamSzegedy, Biggio2013}, in which deliberately designed tiny perturbations can significantly degrade the model performance. Many successful adversarial attacks have been reported in image classification~\cite{FGSM, BIM, AdvPatch,Adv3D}, speech recognition~\cite{Carlini2018}, malware detection~\cite{Grosse2016}, etc.

According to the purpose of the attacker, adversarial attacks can be categorized into two types: non-target attacks and target attacks. In a non-target attack, the attacker wants the model output to an adversarial example (after adding the adversarial perturbation) to be wrong, but does not force it into a particular class. Two typical non-target attack approaches are DeepFool~\cite{Moosavi2016} and universal adversarial perturbation (UAP)~\cite{Moosavi2017}. In a target attack, the model output to an adversarial example should always be biased into a specific wrong class. Some typical target attack approaches are the iterative least-likely class method~\cite{BIM}, adversarial transformation networks~\cite{Baluja2017}, and projected gradient descent (PGD)~\cite{Madry2018}. There are also approaches that can be used in both non-target and target attacks, e.g., L-BFGS~\cite{AdvExamSzegedy}, the fast gradient sign method (FGSM)~\cite{FGSM}, the C\&W method~\cite{AdvCW}, the basic iterative method~\cite{BIM}, etc.

\begin{figure}[htbp]         \centering
\includegraphics[width=.6\linewidth,clip]{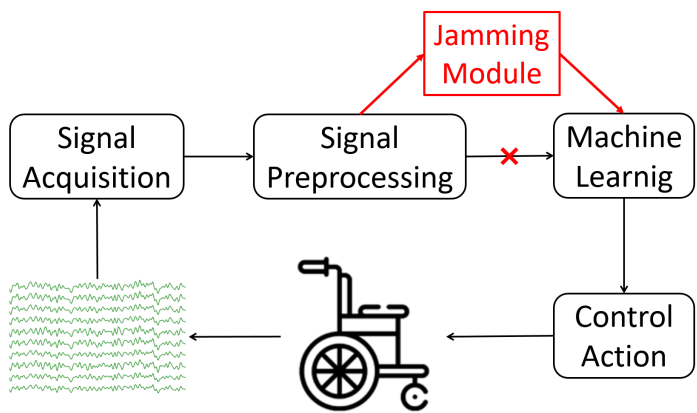}
\caption{Attacking an EEG-based BCI system \cite{Zhang2019}.} \label{fig:BCI}
\end{figure}

An EEG-based BCI system usually consists of four parts: signal acquisition, signal preprocessing, machine learning, and control action. Zhang and Wu~\cite{Zhang2019} explored the vulnerability of CNN classifiers under adversarial attacks in EEG-based BCIs, and discovered that adversarial examples do exist there. They injected a jamming module between signal preprocessing and machine learning to perform the adversarial attack, as shown in Fig.~\ref{fig:BCI}. They successfully attacked three CNN classifiers (EEGNet, DeepCNN, and ShallowCNN) in three different scenarios (white-box, gray-box, and black-box). Their results exposed a critical security problem in EEG-based BCIs, which had not been investigated before. As pointed out in \cite{Zhang2019}, ``\emph{EEG-based BCIs could be used to control wheelchairs or exoskeleton for the disabled \cite{Li2016}, where adversarial attacks could make the wheelchair or exoskeleton malfunction. The consequence could range from merely user confusion and frustration, to significantly reducing the user's quality of life, and even to hurting the user by driving him/her into danger on purpose. In clinical applications of BCIs in awareness evaluation/detection for disorder of consciousness patients \cite{Li2016}, adversarial attacks could lead to misdiagnosis.}"

Albeit their success, Zhang and Wu's approaches \cite{Zhang2019} had the following limitations:
\begin{enumerate}
\item An adversarial perturbation needs to be computed specifically for each input EEG trial, which is inconvenient.
\item To compute the adversarial perturbation, the attacker needs to wait for the complete EEG trial to be collected; however, by that time the EEG trial has gone, and the perturbation cannot be actually added to it. So, Zhang and Wu's approaches are theoretically important, but may not be easily implementable in practice. For better practicability, we should be able to perform the attack as soon as an EEG trial starts.
\end{enumerate}

This paper introduces UAP for BCIs, which is a universal perturbation template computed offline and added to any EEG trial in real-time. Compared with Zhang and Wu's approaches \cite{Zhang2019}, it has two corresponding advantages:
\begin{enumerate}
\item A UAP is computed once and applicable to any EEG trial, instead of being computed specifically for each input EEG trial.
\item A UAP can be added as soon as an EEG trial starts, or anywhere during an EEG trial, thus the attacker does not need to know the number of EEG channels, the starting time, and the length of a trial.
\end{enumerate}
So, it relieves the two limitations of \cite{Zhang2019} simultaneously, making the attack more practical\footnote{A very recent research \cite{drwuNSR2021} also developed adversarial perturbation templates to accommodate casuality, but it considered traditional P300 and SSVEP based BCI speller pipelines, which perform feature extraction and classification separately, whereas in this paper the UAPs are developed for end-to-end deep learning classifiers in P300 and MI tasks.}.

Studies on UAPs appeared in the literature very recently. Moosavi-Dezfooli \emph{et al.}~\cite{Moosavi2017} discovered the existence of UAPs, and verified that they can fool state-of-the-art machine learning models in image classification. Their method for crafting the UAPs, based on DeepFool~\cite{Moosavi2016}, solves a complex optimization problem. The same idea was later used in attacking speech recognition systems~\cite{Neekhara2019}. Behjati \emph{et al.}~\cite{Behjati2019} proposed a gradient projection based approach for generating UAPs in text classification. Mopuri \emph{et al.}~\cite{Mopuri2019} proposed a generalizable and data-free approach for crafting UAPs, which is independent of the underlying task. All these approaches were for non-target attacks. Target attacks using UAPs are more challenging, because the perturbation needs to be both universal and targeting at a particular class. To our knowledge, the only study on UAPs for target attacks was \cite{Hirano2020}, where Hirano \emph{et al.} integrated a simple iterative method for generating non-target UAPs and FGSM for target attacks to generate UAPs for target attacks.

This paper investigates UAPs in EEG-based BCIs. We make the following three contributions:
\begin{enumerate}
\item To our knowledge, this is the first study on UAPs for EEG-based BCIs, which make adversarial attacks in BCIs more convenient and more practical.
\item We propose a novel total loss minimization (TLM) approach for generating a UAP for EEG trials, which can achieve better attack performance with a smaller perturbation, compared with the traditional DeepFool based approach.
\item Our proposed TLM can perform both non-target attacks and target attacks. To our knowledge, no one has studied optimization based UAPs for target attacks before.
\end{enumerate}

The remainder of this paper is organized as follows: Section~\ref{sect:Method} introduces two approaches to generate UAPs for EEG trials. Section~\ref{sect:experiments} describes our experimental setting. Sections~\ref{sect:results} and \ref{sect:results2} present the experimental results on non-target attacks and target attacks, respectively. Finally, Section~\ref{sect:conclusions} draws conclusions and points out several future research directions.

\section{Universal Adversarial Perturbations (UAPs)} \label{sect:Method}

This section first introduces an iterative algorithm for crafting a UAP for EEG trials, then presents the details of our proposed TLM approach. All source code can be downloaded at https://github.com/ZihanLiu95/UAP\_EEG.

We distinguish between two types of attacks:
\begin{itemize}
\item \emph{Target attacks}, in which the attacker wants all adversarial examples to be classified into a \emph{specific} class. For example, for 3-class MI (left-hand, right-hand, and feet), the attacker may want all left-hand and right-hand adversarial trials to be misclassified as feet trials.
\item \emph{Non-target attacks}, in which the attacker wants the adversarial examples to be misclassified, but does not care which class they are classified into. In the above example, a left-hand adversarial trial could be misclassified as a right-hand trial, or a feet trial.
\end{itemize}

\subsection{Problem Setup}

To attack a BCI system, the adversarial perturbations need to be added to benign EEG signals in real-time.

Let $X_i\in\mathbb{R}^{C\times{T}}$ be the $i$-th raw EEG trial $(i=1,...,n$), where $C$ is the number of EEG channels and $T$ the number of time domain samples. Let $\bm{x}\in\mathbb{R}^{C\cdot T\times1}$ be the vector form of $X_i$, which concatenates all columns of $X_i$ into a single column. Let $k(\bm{x}_i)$ be the estimated label from the target CNN model, $\bm{v}\in\mathbb{R}^{C\dot T\times 1}$ be the UAP, and $\tilde{\bm{x}}_i=\bm{x}_i+\bm{v}$ be the adversarial EEG trial after adding the UAP. Then, $\bm{v}$ needs to satisfy:
\begin{eqnarray}
\left.
\begin{array}{rcl}
  \frac{1}{n}\sum_{i=1}^{n}I\left(k(\bm{x}_i+\bm{v})\neq{k(\bm{x}_i)}\right) & \geq& \delta \\
  \|\bm{v}\|_{p} & \leq &\xi
\end{array}\right\}, \label{eq:cons}
\end{eqnarray}
where $\|\cdot\|_p$ is the $L_p$ norm, and $I(\cdot)$ is the indicator function which equals $1$ if its argument is true, and $0$ otherwise. The parameter $\delta\in(0,1]$ determines the desired attack success rate (ASR), and $\xi$ constrains the magnitude of $\bm{v}$. Briefly speaking, the first constraint requires the UAP to achieve a desired ASR, and the second constraint ensures the UAP is small.

Next, we describe how a UAP can be crafted for EEG data. We first introduce DeepFool~\cite{Moosavi2016}, a white-box attack (the attacker has access to all information of the victim model, including its architecture, parameters, and training data) approach for crafting an adversarial perturbation for a \emph{single} input example, and then extend it to crafting a UAP for \emph{multiple} examples. Finally, we propose a novel TLM approach to craft UAP, which can be applied to both non-target attacks and target attacks.

\subsection{DeepFool-Based UAP}

DeepFool is an approach for crafting an adversarial perturbation for a single input example.

Consider a binary classification problem, where the labels are $\{-1,1\}$. Let $\bm{x}$ be an input example, and $f$ an affine classification function $f(\bm{x})=\bm{w}^{T}\bm{x}+b$. Then, the predicted label is $k(\bm{x})={\rm sign}f(\bm{x})$. The minimal adversarial perturbation $\bm{r}^{*}$ should move $\bm{x}$ to the decision hyperplane $\mathcal{F}=\{\bm{x}^*:\bm{w}^{T}\bm{x}^*+b=0\}$, i.e.,
\begin{eqnarray}
\bm{r}^{*}=-\frac{f(\bm{x})}{\|\bm{w}\|^{2}_{2}}\bm{w}. \label{eq:r}
\end{eqnarray}
%which is equivalent to:
%\begin{align}
%f(\bm{x})+\frac{\|\bm{w}\|^{2}_{2}}{\bm{w}}\bm{r}^{*}=0.
%\end{align}

CNN classifiers are nonlinear. So, an iterative procedure \cite{Moosavi2016} is used to identify the adversarial perturbation, by approximately linearizing $f(\bm{x}_t) \approx f(\bm{x}_t)+\nabla{f(\bm{x}_t)}^{T}\bm{r}_t$ around $\bm{x}_t$ at Iteration $t$, where $\nabla$ is the gradient of $f(\bm{x}_t)$. Then, the minimal perturbation at Iteration $t$ is computed as:
\begin{eqnarray}
\min\limits_{\bm{r}_t}\ \|\bm{r}_t\|,\quad \mbox{s.t.} \ f(\bm{x}_t)+\nabla{f(\bm{x}_t)}^{T}\bm{r}_t=0. \label{eq:deepfool}
\end{eqnarray}

The perturbation $\bm{r}_t$ at Iteration $t$ is computed using the closed-form solution in (\ref{eq:r}), and then $\bm{x}_{t+1}=\bm{x}_t+\bm{r}_t$ is used in the next iteration. The iteration stops when $\bm{x}_{t+1}$ starts to change the classification label. The pseudocode is given in Algorithm~\ref{alg:Deepfool}.

\begin{algorithm}[htpb] %\DontPrintSemicolon
\KwIn{$\bm{x}$, an input example\;
\hspace*{11mm}$f$, the classification function.}
\KwOut{$\bm{r}^{*}$, the adversarial perturbation.}
$\bm{x}_0=\bm{x}$\;
$t=0$\;
\While{${\rm sign}f(\bm{x}_t)={\rm sign}f(\bm{x}_0)$}{
$\bm{r}_t=-\frac{f(\bm{x}_t)}{\|\nabla{f(\bm{x}_t)}\|^{2}_{2}}\nabla{f(\bm{x}_t)}$\;
$\bm{x}_{t+1}=\bm{x}_t+\bm{r}_t$\;
$t={t+1}$\;
}
$\bm{r}^{*}=\sum_{i=0}^t{\bm{r}_i}$.
\caption{DeepFool \cite{Moosavi2016} for generating an adversarial perturbation for a \emph{single} input example.} \label{alg:Deepfool}
\end{algorithm}

Algorithm~\ref{alg:Deepfool} can be extended to multi-class classification by using the one-versus-all scheme to find the closest hyperplane. Experiments in \cite{Moosavi2016} demonstrated that DeepFool can achieve comparable attack performance as FGSM~\cite{FGSM}, but the magnitude of the perturbation is smaller, which is more desirable.

UAPs were recently discovered in image classification by Moosavi-Dezfooli \emph{et al.}~\cite{Moosavi2017}, who showed that a fixed adversarial perturbation can fool multiple state-of-the-art CNN classifiers on multiple images. They developed a DeepFool-based iterative algorithm to craft the UAP, which satisfies (\ref{eq:cons}). A UAP is designed by proceeding iteratively over all examples in the dataset $\bm{X}=\{\bm{x}_i\}_{i=1}^n$. In each iteration, DeepFool is used to compute a minimum perturbation $\bigtriangleup{\bm{v}_i}$ for the current perturbed point $\bm{x}_i+\bm{v}$, and then $\bigtriangleup{\bm{v}_i}$ is aggregated into $\bm{v}$.

More specifically, if the current universal perturbation $\bm{v}$ cannot fool the classifier on $\bm{x}_i$, then a minimum extra perturbation $\bigtriangleup{\bm{v}_i}$ that can fool $\bm{x}_i$ is computed by solving the following optimization problem:
\begin{eqnarray}
\min_{\bigtriangleup{\bm{v}_i}}\|\bigtriangleup{\bm{v}_i}\|_2, \quad \mbox{s.t.}\ \ k(\bm{x}_i+\bm{v}+\bigtriangleup{\bm{v}_i})\neq k(\bm{x}_i). \label{eq:dv}
\end{eqnarray}

To ensure the constraint $\|\bm{v}\|_{p}\leq\xi$ is satisfied, the updated universal perturbation $\bm{v}$ is further projected onto the $\ell_p$ ball of radius $\xi$ centered at 0. The projection operator $\mathcal{P}_{p,\xi}$ is defined as:
\begin{eqnarray}
\mathcal{P}_{p,\xi}(\bm{v})=\mathrm{arg}\min\limits_{\|\bm{v}'\|_p\leq\xi}\|\bm{v}-\bm{v}'\|_{2}. \label{eq:P}
\end{eqnarray}

Then, the UAP can be updated by $\bm{v}=\mathcal{P}_{p,\xi}(\bm{v}+\bigtriangleup{\bm{v}_i})$ in each iteration. This process is repeated on the entire dataset until the maximum number of iterations is reached, or the ASR on the perturbed dataset $\bm{X}_v=\{\bm{x}_i + \bm{v}\}_{i=1}^{n}$ exceeds the target ASR threshold $\delta\in(0,1]$, i.e.,
\begin{eqnarray}
ASR(\bm{X}_v,\bm{X})=\frac{1}{n}{\sum_{i=1}^{n}I\left(k(\bm{x_i}+\bm{v})\neq{k(\bm{x_i})}\right)\geq \delta}.
\end{eqnarray}

The pseudo-code of the DeepFool-based algorithm is given in Algorithm~\ref{alg:UAP}.

\begin{algorithm}[ht] %\DontPrintSemicolon
\KwIn{$\bm{X}=\{\bm{x}_i\}_{i=1}^{n}$, $n$ input examples\;
\hspace*{11mm}$k$, the classifier\;
\hspace*{11mm}$\xi$, the maximum $\ell_p$ norm of the UAP\;
\hspace*{11mm}$\delta$, the desired ASR\;
\hspace*{11mm}$M$, the maximum number of iterations.}
\KwOut{$\bm{v}$, a UAP.}

$\bm{v}=\bm{0}$\;
$\bm{X}_v=\bm{X}$\;
\For{$m=1,...,M$}{
\uIf{$ASR(\bm{X}_v,\bm{X})<\delta$}{

\For{Each $\bm{x}_i\in{\bm{X}}$}{
\If{$k(\bm{x}_i+\bm{v})==k(\bm{x}_i)$}{
Use DeepFool to compute the minimal perturbation $\bigtriangleup{\bm{v}_i}$ in (\ref{eq:dv});\

Update the perturbation by (\ref{eq:P}): $\bm{v}\gets{\mathcal{P}_{p,\xi}}(\bm{v}+\bigtriangleup{\bm{v}_i})$;\
}
}
$\bm{X}_v=\{\bm{x}_i + \bm{v}\}_{i=1}^{n}$;
}
\Else{
Break;\
}
}
\textbf{Return} $\bm{v}$.
\caption{DeepFool-based algorithm for generating a UAP \cite{Moosavi2017}.} \label{alg:UAP}
\end{algorithm}

\subsection{Our Proposed TLM-Based UAP}

Different from the DeepFool-based algorithm, TLM directly optimizes an objective function w.r.t. the UAP by batch gradient descent. In white-box attacks, the parameters of the victim model are known and fixed, and hence we can view the UAP as a variable to minimize an objective function on the entire training set.

Specifically, we solve the following optimization problem by gradient descent:
\begin{eqnarray}
\min_{\bm{v}} E_{\bm{x}\sim{\bm{D}}}l(\bm{x}+\bm{v},y)+\alpha\cdot{C(\bm{x},\bm{v})},\quad\mbox{s.t.} \|\bm{v}\|_{p} \leq \xi,\label{eq:expectation}
\end{eqnarray}
where $l(\bm{x}+\bm{v},y)$ is a loss function, which evaluates the effect of the UAP on the target model. $C(\bm{x},\bm{v})$ is the constraint on the perturbation $\bm{v}$, and $\alpha$ the regularization coefficient.
Our proposed approach is highly flexible, as the attacker can choose different optimizers, loss functions, and/or constraints, according to the specific task.

For non-target attacks, the loss function $l$ can be defined as $l(\bm{x},y)=\log(p_{y}(\bm{x}))$, in which $y$ is the true label of $\bm{x}$. This loss function forces the UAP to make the model to have minimum confidence on the true label $y$. TLM optimizes the perturbation by increasing the expected loss of the model on the training data $\bm{D}$ as much as possible. When UAP can affect enough samples in the dataset, and the distribution difference between the test data and $\bm{D}$ is small, UAP will have a high probability of affecting the test data. In practice, we could also use $\arg\max\limits_{j}p_{j}(\bm{x})$, i.e., the predicted label, to replace $y$ if the true label is not available.

For target attacks, the loss function should force the UAP to maximize the model's confidence on the target label $y_t$ specified by attacker; hence, $l$ can be defined as $l(\bm{x},y_t)=-\log(p_{y_t}(\bm{x}))$. In fact, both $l(\bm{x},y)$ and $l(\bm{x},y_t)$ are essentially the negative cross-entropy.

There are also various options for the constraint function $C(\bm{x},\bm{v})$. In most cases, we can simply set $C(\bm{x},\bm{v})$ as L1 or L2 regularization on the UAP $\bm{v}$; however, it can also be a more sophisticated function, e.g., a metric function to detect whether the input is an adversarial example or not. When a new metric function for detecting adversarial examples is proposed, our approach can also be utilized to test its reliability: we set $C$ as the metric function to check whether we can still find an adversarial example. Given the diversity of metric functions, we only consider L1 or L2 regularization in this paper. Other metric functions and defense strategies for TLM-UAP will be considered in our future research.

To ensure the constraint $\|\bm{v}\|_{p} \leq \xi$, we can project the UAP $\bm{v}$ into the $l_p$ ball of radius $\xi$ centered at $0$ by the projection function in (\ref{eq:P}) after each optimization iteration, or simply clip its amplitude into $[-\xi,\xi]$, which is equivalent to using the $l_\infty$ ball. The latter was used in this paper for its simplicity.

Due to the transferability of adversarial examples, i.e., adversarial examples generated by one model may also be used to attack another one, we can perform TLM-UAP attacks in a gray-box attack scenario. In this case, the attacker only has access to the training set of the victim model, instead of its architecture and parameters. The attacker can train a substitute model on the known training set to generate a UAP, which can then be used to attack the victim model.

The TLM-UAP can also be simplified, e.g., the same perturbation is designed for all EEG channels, or a mini TLM-UAP is added to an arbitrary location of an EEG trail. The corresponding experimental results are shown in Sections~\ref{sect:White-Box-results} and \ref{sect:results2}. %This scaled template could be implemented more effective when we have limited access to the information of EEG trails.

The pseudo-code of our proposed TLM approach is given in Algorithm~\ref{alg:TLM-UAP}.

\begin{algorithm}[htpb] %\DontPrintSemicolon
\KwIn{$\bm{X}_{train}=\{\bm{x}_{train,i}\}_{i=1}^{n}$, $n$ training examples\;
\hspace*{11mm}$\bm{X}_{val}=\{\bm{x}_{val,i}\}_{i=1}^{m}$, $m$ validation examples\;
\hspace*{11mm}$k$, the classifier\;
\hspace*{11mm}$\xi$, the maximum $\ell_p$ norm of the UAP\;
\hspace*{11mm}$\alpha$, the regularization coefficient\;
\hspace*{11mm}$\delta$, the desired ASR\;
\hspace*{11mm}$M$, the maximum number of epochs\;}
\KwOut{$\bm{v}_{best}$, a UAP.}
$\bm{v}=\bm{0}$\;
$r=0$\;
\For{$m=1,...,M$}{
    \For{Each mini-batch $\bm{D}\in{\bm{X}_{train}}$}{
        Update $\bm{v}$ in (\ref{eq:expectation}) for $\bm{D}$ with an optimizer\;
        Constrain $\bm{v}$ by (\ref{eq:P}): $\bm{v}\gets{\mathcal{P}_{p,\xi}}(\bm{v})$, or directly clip $\bm{v}$ into $[-\xi,\xi]$;\
    }
    $\bm{X}_{val,\bm{v}}=\{\bm{x}_{val,i} + \bm{v}\}_{i=1}^{n}$;

    \If{$ASR(\bm{X}_{val,\bm{v}},\bm{X}_{val})>r$}{
        $r=ASR(\bm{X}_{val,\bm{v}},\bm{X}_{val})$\;
        $\bm{v}_{best}=\bm{v}$;\
    }
    \If{$r>\delta$}{
        Break;\
    }
}
\textbf{Return} $\bm{v}_{best}$.
\caption{The proposed TLM approach for generating a UAP.} \label{alg:TLM-UAP}
\end{algorithm}

\section{Experimental Settings} \label{sect:experiments}

This section introduces the experimental settings for validating the performance of our proposed TLM approach.

\subsection{The Three BCI Datasets}

The following three BCI datasets were used in our experiments, as in \cite{drwuALBCI2019,Zhang2019}:

\textbf{P300 evoked potentials (P300)}: The P300 dataset, first introduced in \cite{EPFLP300}, contained eight subjects. In the experiment, each subject faced a laptop on which six images were flashed randomly to elicit P300 responses. The goal was to classify whether the image was a target or non-target. The 32-channel EEG data was downsampled to 256Hz, bandpass filtered to $[1,40]$ Hz, and epoched to $[0,1]$s after each image onset. Then, we normalized the data using $\frac{x-\footnotesize{\mbox{mean}}(x)}{10}$, and clipped the resulting values to $[-5, 5]$. Each subject had about 3,300 trials.

\textbf{Feedback error-related negativity (ERN)}: The ERN dataset \cite{ERN} was used in a Kaggle challenge\footnote{https://www.kaggle.com/c/inria-bci-challenge}. The EEG signals were collected from 26 subjects and consisted of two classes (bad-feedback and good-feedback). The entire dataset was partitioned into a training set (16 subjects) and a test set (10 subjects). We used all 26 subjects in the experiments. The 56-channel EEG signals were downsampled to 200Hz, bandpass filtered to $[1,40]$Hz, epoched to $[0,1.3]$s after each stimulus, and $z$-normalized. Each subject had 340 trials.

\textbf{Motor imagery (MI)}: The MI dataset was Dataset 2A\footnote{http://www.bbci.de/competition/iv/} in BCI Competition IV~\cite{MI4C}. The EEG signals were collected from nine subjects and consisted of four classes: the imagined movements of the left hand, right hand, both feet, and tongue. The 22-channel EEG signals were downsampled to 128Hz, bandpass filtered to $[4,40]$Hz, epoched to $[0,2]$s after each imagination prompt, and standardized using an exponential moving average window with a decay factor of 0.999, as in \cite{EEGNet}. Each subject had 576 trials, with 144 in each class.

\subsection{The Three CNN Models}

The following three CNN models were used in our experiments, as in \cite{drwuALBCI2019,Zhang2019}:

\textbf{EEGNet}: EEGNet~\cite{EEGNet} is a compact CNN architecture for EEG-based BCIs. It consists of two convolutional blocks and a classification block. To reduce the number of model parameters, EEGNet uses depthwise and separable convolutions~\cite{Xception} instead of traditional convolutions.

\textbf{DeepCNN}: DeepCNN \cite{MNE} consists of four convolutional blocks and a softmax layer for classification, which is deeper than EEGNet. Its first convolutional block is specially designed to handle EEG inputs, and the other three are standard convolutional blocks.

\textbf{ShallowCNN}: Inspired by filter bank common spatial patterns~\cite{FBCSP}, ShallowCNN \cite{MNE} is specifically tailored to decode band power features. Compared with DeepCNN, ShallowCNN uses a larger kernel in temporal convolution, and then a spatial filter, squaring nonlinearity, a mean pooling layer and logarithmic activation function.

\subsection{The Two Experimental Settings}

We considered two experimental settings:

\textbf{Within-subject experiments}: Within-subject 5-fold cross-validation was used in the experiments. For each individual subject, all EEG trials were divided into 5 non-overlapping blocks. Three blocks were selected for training, one for validation, and the remaining one for test. We made sure each block was used in test once, and reported the average results.

\textbf{Cross-subject experiments}: For each dataset, leave-one-subject-out cross-validation was performed. Assume a dataset had $N$ subjects, and the $N$-th subject was selected as the test subject. In training, trials from the first $N-1$ subjects were mixed, and divided into 75\% for training and 25\% for validation in early stopping.

When training the victim models on the first two datasets, we applied weights to different classes to accommodate the class imbalance, according to the inverse of its proportion in the training set. We used the cross entropy loss function and Adam optimizer~\cite{Adam}. Early stopping was used to reduce overfitting.

The parameters for generating DF-UAP and TLM-UAP are shown in Table~\ref{tab:parameters}. It should be noted that TLM-UAP was trained with no constraint, and we set $\delta$ to $1.0$ and used early stopping (patience=10) to decide whether to stop the iteration or not. We replaced the true labels $y$ in (\ref{eq:expectation}) with the predicted ones, as in real-world applications we do not have the true labels.

\begin{table}[htbp] \center
\caption{Parameters for generating DF-UAP and TLM-UAP. $\|\bm{v}\|_{\infty}$ was used in computing the norm of the UAPs.}   \label{tab:parameters}
\begin{indented}
\item[]\begin{tabular}{c|ccccc}
\toprule
            & $\xi$           & $\delta$  & $M$   & $\alpha$  & Constraint \\ \hline
     DF-UAP     & 0.2         & 0.8       & 10    & -         & -     \\
     TLM-UAP    & 0.2        & 1.0       & 500    & 0       & No    \\
\bottomrule
\end{tabular}
\end{indented}
\end{table}

\subsection{The Two Performance Measures}

Both raw classification accuracy (RCA) and balanced classification accuracy (BCA) \cite{drwuTHMS2017} were used as the performance measures. The RCA is the ratio of the number of correctly classified samples to the number of total samples, and the BCA is the average of the individual RCAs of different classes.

The BCA is necessary, because some BCI paradigms (e.g., P300) have intrinsic significant class imbalance, and hence using RCA alone may be misleading sometimes.

\section{Non-Target Attack Results} \label{sect:results}

This section presents the experimental results in non-target attacks on the three BCI datasets. Recall that a non-target attack forces a model to misclassify an adversarial example to any class, instead of a specific class.

For notation convenience, we denote the UAP generated by the DeepFool-based algorithm (Algorithm~\ref{alg:UAP}) as \emph{DF-UAP}, and the UAP generated by the proposed TLM (Algorithm~\ref{alg:TLM-UAP}) as \emph{TLM-UAP}.

\subsection{Baseline Performances}

We compared the UAP attack performance with two baselines:

\subsubsection{Clean Baseline}

We evaluated the baseline performances of the three CNN models on the clean (unperturbed) EEG data, as shown in the first part of Table~\ref{tab:results}. For all three datasets and all three classifiers, generally RCAs and BCAs of the within-subject experiments were higher than their counterparts in the cross-subject experiments, which is reasonable, since individual differences cause inconsistency among EEG trials from different subjects.

\begin{table*}[htbp] \center
\caption{The ratio of the number of correctly classified samples to the number of total samples (RCAs), and the mean RCAs of different classes (BCAs), of the three CNN classifiers in different non-target attack scenarios on the three datasets ($\xi=0.2$). For each attack type on each CNN model and each dataset, the best performances are marked in bold. Statistically significantly different RCAs/BCAs between DF-UAP and TLM-UAP were marked with `*' (non-parametric Mann-Whitney \emph{U} test; $p<0.01$).}  \setlength{\tabcolsep}{0.5mm}  \label{tab:results}
\scalebox{0.65}{
\begin{tabular}{c|c|c|c|c|c|c|ccc|ccc} \toprule
\multirow{4}{*}{Experiment}&\multirow{4}{*}{Dataset}&\multirow{4}{*}{Victim Model}  &\multicolumn{10}{c}{RCA/BCA}\\ \cline{4-13}
&&&\multicolumn{2}{c|}{Baseline} & \multicolumn{2}{c|}{White-Box Attack} &\multicolumn{6}{c}{Gray-Box Attack}\\ \cline{4-13}
& & &\multirow{2}{*}{Clean} & \multirow{2}{*}{Noisy}& \multirow{2}{*}{DF-UAP} & \multirow{2}{*}{TLM-UAP} & \multicolumn{3}{|c|}{Substitute Model (DF-UAP)}   & \multicolumn{3}{|c}{Substitute Model (TLM-UAP)}       \\  \cline{8-10}  \cline{11-13}
 &           &            &        &    &   &    &EEGNet         &DeepCNN        &ShallowCNN    &EEGNet         &DeepCNN      &ShallowCNN\\
\midrule
&\multirow{3}{*}{P300}  &EEGNet      &$.81/.79$ &$.80/.79$ &$.18/.51$    &$\mathbf{.17^*/.50^*}$ &$.21/.52$  &$.21/.52$ &$.40/.62$ &$\mathbf{.17^*/.20^*}$ &$\mathbf{.19^*/.50^*}$ &$\mathbf{.25^*/.54^*}$  \\
                     &  &DeepCNN     &$.82/.78$ &$.82/.78$ &$.20/.51$    &$\mathbf{.19^*/.50^*}$ &$.33/.58$  &$.24/.53$ &$.49/.65$ &$\mathbf{.20^*/.51^*}$ &$\mathbf{.20^*/.51^*}$ &$\mathbf{.30^*/.57^*}$  \\
                     &  &ShallowCNN  &$.80/.75$ &$.80/.74$ &$.46/.62$    &$\mathbf{.20^*/.51^*}$ &$.62/.69$  &$.56/.67$ &$.58/.66$ &$\mathbf{.44^*/.62^*}$ &$\mathbf{.43^*/.61^*}$ &$\mathbf{.29^*/.54^*}$ \\ \cline{2-13}

Within &\multirow{3}{*}{ERN}&EEGNet  &$.76/.73$ &$.69/.64$ &$.32/.51$          &$\mathbf{.30/.50}$  &$.58/.64$  &$.56/.63$ &$.69/.67$  &$\mathbf{.45^*/.57^*}$  &$\mathbf{.53/.61}$  &$\mathbf{.62^*/.66^*}$ \\
-Subject             &  &DeepCNN     &$.69/.65$ &$.69/.65$ &$.44/.55$          &$\mathbf{.40/.52}$ &$.65/.64$  &$.60/.62$ &$.66/.62$   &$\mathbf{.60^*/.62^*}$  &$\mathbf{.59/.61}$  &$\mathbf{.64/.62}$ \\
                     &  &ShallowCNN  &$.70/.68$ &$.69/.67$ &$.52/.57$          &$\mathbf{.40^*/.53^*}$  &$\mathbf{.66/.65}$  &$.65/.65$ &$.62/.62$  &$.66/.66$  &$\mathbf{.64/.65}$  &$\mathbf{.57^*/.61^*}$                            \\ \cline{2-13}

&\multirow{3}{*}{MI}    &EEGNet      &$.61/.61$ &$.47/.48$ &$.29/.29$          &$\mathbf{.25^*/.25^*}$  &$.37/.38$  &$.36/.36$ &$.38/.38$ &$\mathbf{.34/.34}$  &$\mathbf{.31^*/.31^*}$  &$\mathbf{.34/.34}$  \\
                     &  &DeepCNN     &$.50/.50$ &$.47/.47$ &$.35/.35$          &$\mathbf{.26^*/.26^*}$  &$\mathbf{.47/.47}$  &$.41/.41$ &$.44/.44$ &$.49/.49$  &$\mathbf{.32^*/.32^*}$  &$\mathbf{.39/.39}$  \\
                     &  &ShallowCNN  &$.74/.74$ &$.68/.68$ &$.29/.29$          &$\mathbf{.25^*/.25^*}$  &$\mathbf{.55^*/.55^*}$  &$.40/.40$ &$.32/.32$ &$.64/.63$  &$\mathbf{.35^*/.35^*}$  &$\mathbf{.27^*/.27^*}$  \\
\bottomrule
  &\multirow{3}{*}{P300}&EEGNet      &$.68/.63$ &$.69/.63$ &$.19/.51$          &$\mathbf{.17^*/.50^*}$  &$.22/.52$  &$.29/.54$ &$.27/.54$ &$\mathbf{.17^*/.50^*}$  &$\mathbf{.17^*/.50^*}$  &$\mathbf{.18^*/.50^*}$  \\
                     &  &DeepCNN     &$.69/.64$ &$.70/.64$ &$.20/.51$          &$\mathbf{.18^*/.50^*}$  &$.24/.52$  &$.22/.52$ &$.28/.54$ &$\mathbf{.18^*/.50^*}$  &$\mathbf{.18^*/.50^*}$  &$\mathbf{.18^*/.50^*}$  \\
                     &  &ShallowCNN  &$.67/.62$ &$.66/.62$ &$.27/.54$          &$\mathbf{.19^*/.50^*}$  &$.32/.55$  &$.35/.57$ &$.32/.55$ &$\mathbf{.20^*/.51^*}$  &$\mathbf{.20^*/.51^*}$  &$\mathbf{.18^*/.50^*}$  \\ \cline{2-13}

Cross &\multirow{3}{*}{ERN} &EEGNet  &$.67/.68$ &$.67/.65$ &$.31/.51$          &$\mathbf{.29/.50^*}$  &$.53/.59$  &$.60/.58$ &$.43/.59$ &$\mathbf{.29^*/.50^*}$  &$\mathbf{.31^*/.50^*}$  &$\mathbf{.29^*/.50^*}$   \\
-Subject             &  &DeepCNN     &$.69/.69$ &$.71/.70$ &$.34/.52$          &$\mathbf{.31/.50}$    &$.49/.54$  &$.41/.53$ &$.36/.53$ &$\mathbf{.29^*/.50^*}$  &$\mathbf{.31^*/.50^*}$  &$\mathbf{.29^*/.50^*}$   \\
                     &  &ShallowCNN  &$.69/.69$ &$.68/.68$ &$.38/.55$          &$\mathbf{.29^*/.50^*}$ &$.60/.63$  &$.65/.62$ &$.42/.56$ &$\mathbf{.35^*/.54^*}$  &$\mathbf{.38^*/.55^*}$  &$\mathbf{.29^*/.50^*}$   \\ \cline{2-13}

  &\multirow{3}{*}{MI}  &EEGNet      &$.44/.44$ &$.38/.38$ &$.30/.30$          &$\mathbf{.25^*/.25^*}$  &$.30/.30$  &$.34/.34$ &$.30/.30$ &$\mathbf{.29/.29}$   &$\mathbf{.29^*/.29^*}$ &$\mathbf{.25^*/.25^*}$   \\
                     &  &DeepCNN     &$.47/.47$ &$.44/.44$ &$.33/.33$          &$\mathbf{.25^*/.25^*}$  &$\mathbf{.38/.38}$  &$.32/.32$ &$.28/.28$ &$.40/.40$   &$\mathbf{.30/.30}$  &$\mathbf{.26/.26}$   \\
                     &  &ShallowCNN  &$.47/.47$ &$.43/.43$ &$.27/.27$          &$\mathbf{.25^*/.25^*}$  &$\mathbf{.31^*/.31^*}$  &$\mathbf{.26/.26}$ &$.30/.30$ &$.35/.35$  &$.29/.29$  &$\mathbf{.26^*/.26^*}$   \\
\bottomrule
\end{tabular} }
\end{table*}

\subsubsection{Noisy Baseline}

We added clipped Gaussian random noise $\xi\cdot\max(-1,\min(1,\mathcal{N}(0,1)))$ to the original EEG data, which had the same maximum amplitude $\xi$ as the UAP and satisfied the constraint $\|\bm{v}\|_{p} \leq \xi$ after clipping. If the random noise under the same magnitude constraint can significantly degrade the classification performance, then there is no need to compute a UAP. The results are shown in Table~\ref{tab:results}. Random noise with the same amplitude as the UAP did not degrade the classification performance in most cases, except sometimes on the MI dataset. This suggests that the three CNN classifiers are generally robust to random noise, therefore we should deliberately design the adversarial perturbations.

\subsection{White-Box Non-target Attack Performances}  \label{sect:White-Box-results}

First consider white-box attacks, where we have access to all information of the victim model, including its architecture, parameters, and training data. The performances of DF-UAP and TLM-UAP in white-box non-target attacks are shown in the second part of Table~\ref{tab:results}. We also performed non-parametric Mann-Whitney \emph{U} tests \cite{Mann1947} on the RCAs (BCAs) of DF-UAP and TLM-UAP to check if there were statistically significant differences between them (marked with `*'; $p<0.01$). Observe that:
\begin{enumerate}
\item After adding DF-UAP or TLM-UAP, both the RCAs and the BCAs were significantly reduced, suggesting the effectiveness of DF-UAP and TLM-UAP attacks.
\item In most cases, TLM-UAP significantly outperformed DF-UAP. This may be due to the fact that, as shown in (7), TLM-UAP optimizes the ASR directly on the entire dataset, whereas DF-UAP considers each sample individually, which may be easily trapped into a local minimum.
\item The BCAs of the P300 and ERN datasets were close to $0.5$ after DF-UAP or TLM-UAP attacks, whereas the RCAs were lower than $0.5$, implying that most test EEG trials were classified into the minority class to achieve the best attack performance.
\end{enumerate}

\begin{figure*}[htpb]\centering
\subfigure[]{ \label{fig:P300_label}   \includegraphics[width=.32\columnwidth,clip]{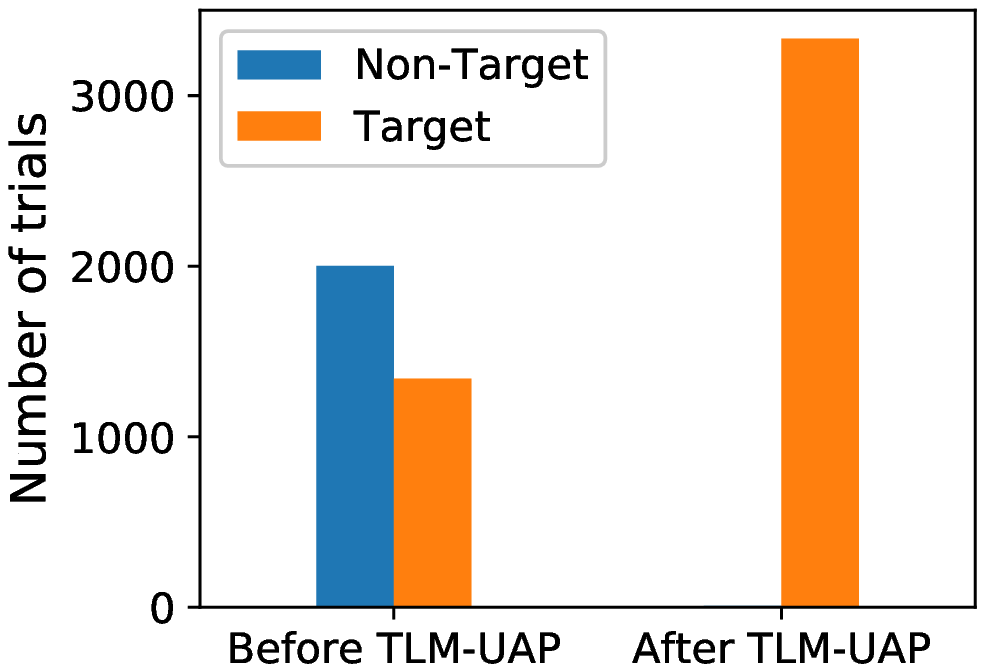}}
\subfigure[]{ \label{fig:ERN_label}    \includegraphics[width=.32\columnwidth,clip]{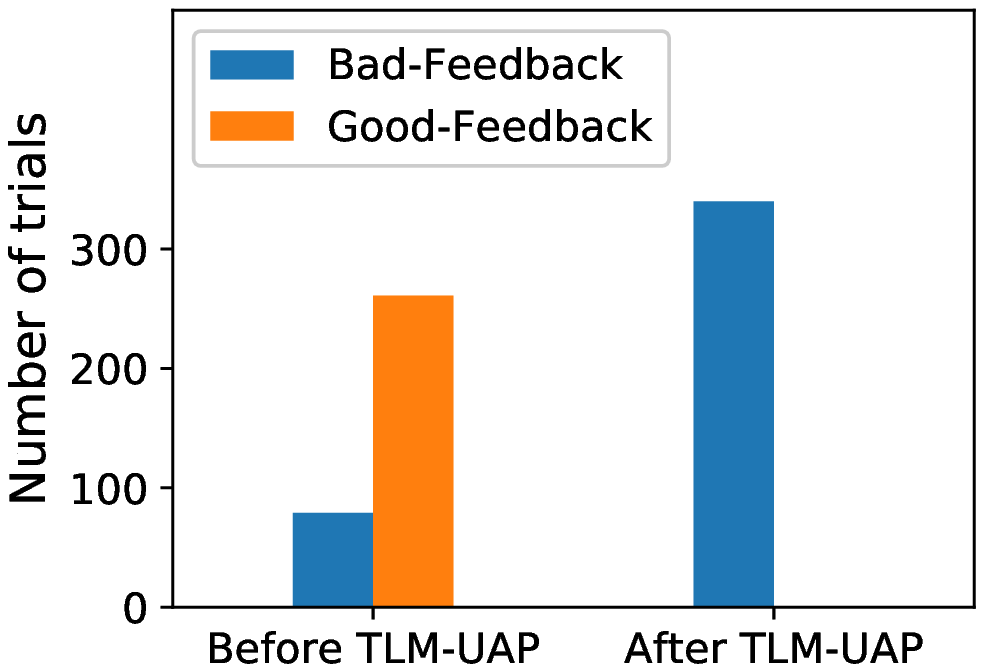}}
\subfigure[]{ \label{fig:MI_label}     \includegraphics[width=.32\columnwidth,clip]{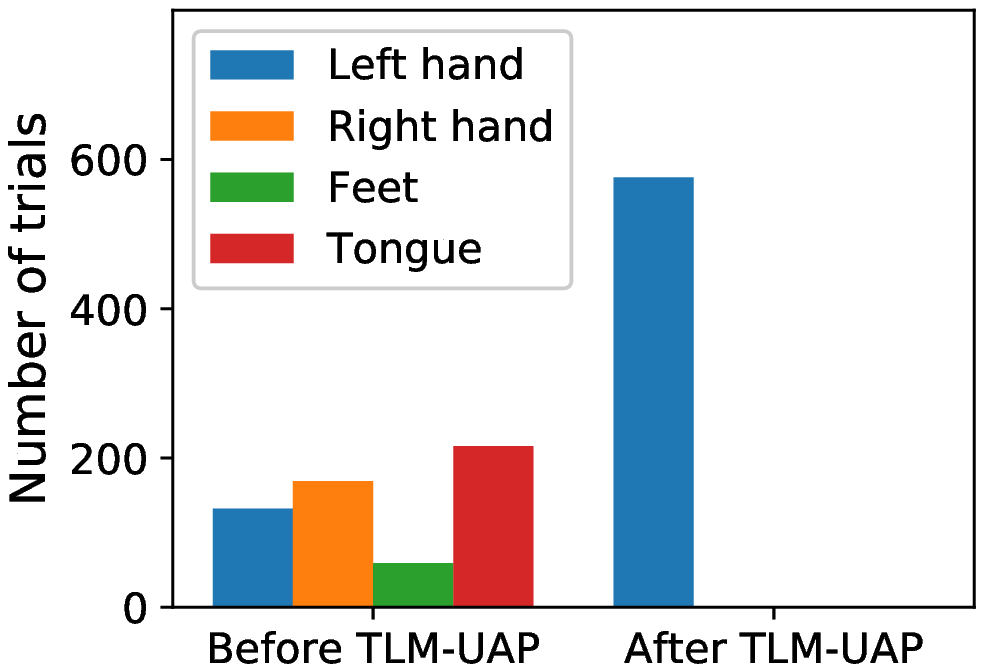}}
\caption{Number of EEG trials in each class (classified by EEGNet), before and after applying TLM-UAP in white-box non-target attack. a) P300; b) ERN; and, c) MI.} \label{fig:label}
\end{figure*}

Fig.~\ref{fig:label} shows the number of EEG trials in each class on the three datasets, classified by EEGNet before and after applying TLM-UAP. Generally, the trials originally classified into the majority class were misclassified into the minority class in binary classification after applying TLM-UAP. This is reasonable. Assume the minority class contains $p\%$ ($p<50$) of the trials. Then, misclassifying all minority class trials into the majority class gives an ASR of $p\%$, whereas misclassifying all majority class trials into the minority class gives an ASR of $(100-p)\%$. Clearly, the latter is larger. Similarly, the trials were misclassified into a minority class (but not necessarily the smallest class) in multi-class classification.

\begin{figure}[htpb]   \centering
\subfigure[]{ \label{fig:time_example}    \includegraphics[width=.9\columnwidth,clip]{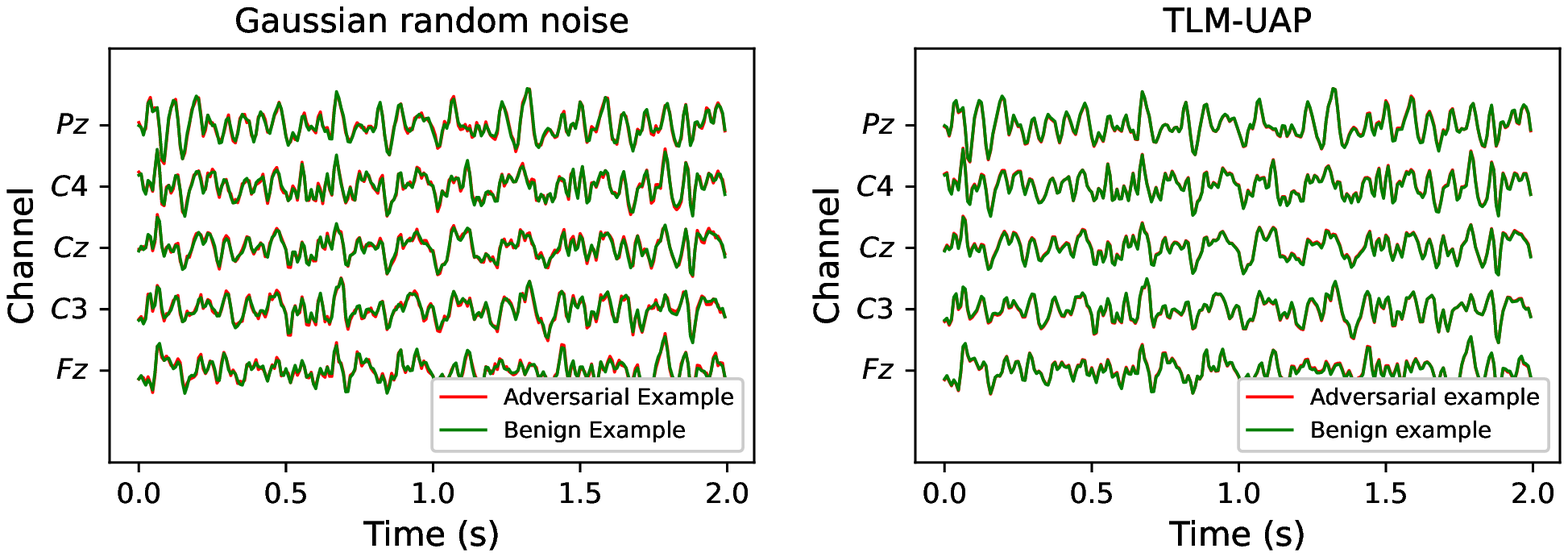}}
\subfigure[]{ \label{fig:fre_example}     \includegraphics[width=.9\columnwidth,clip]{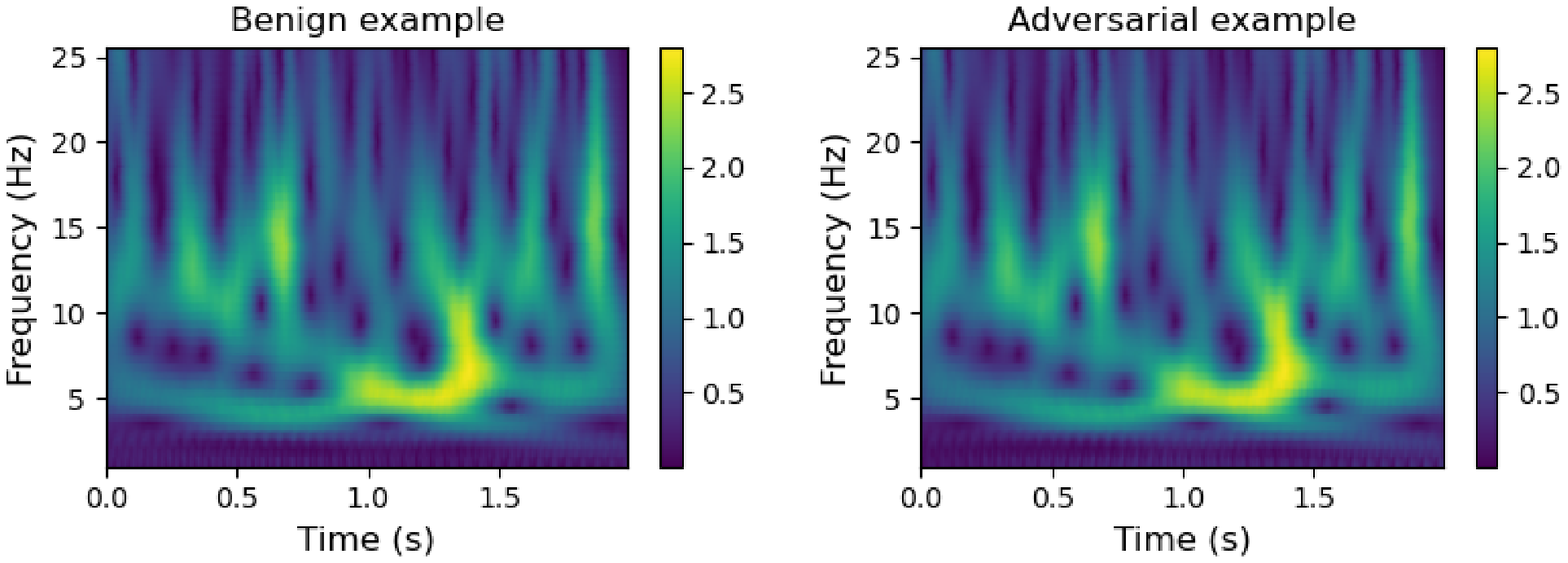}}
\subfigure[]{ \label{fig:topomap_example} \includegraphics[width=.9\columnwidth,clip]{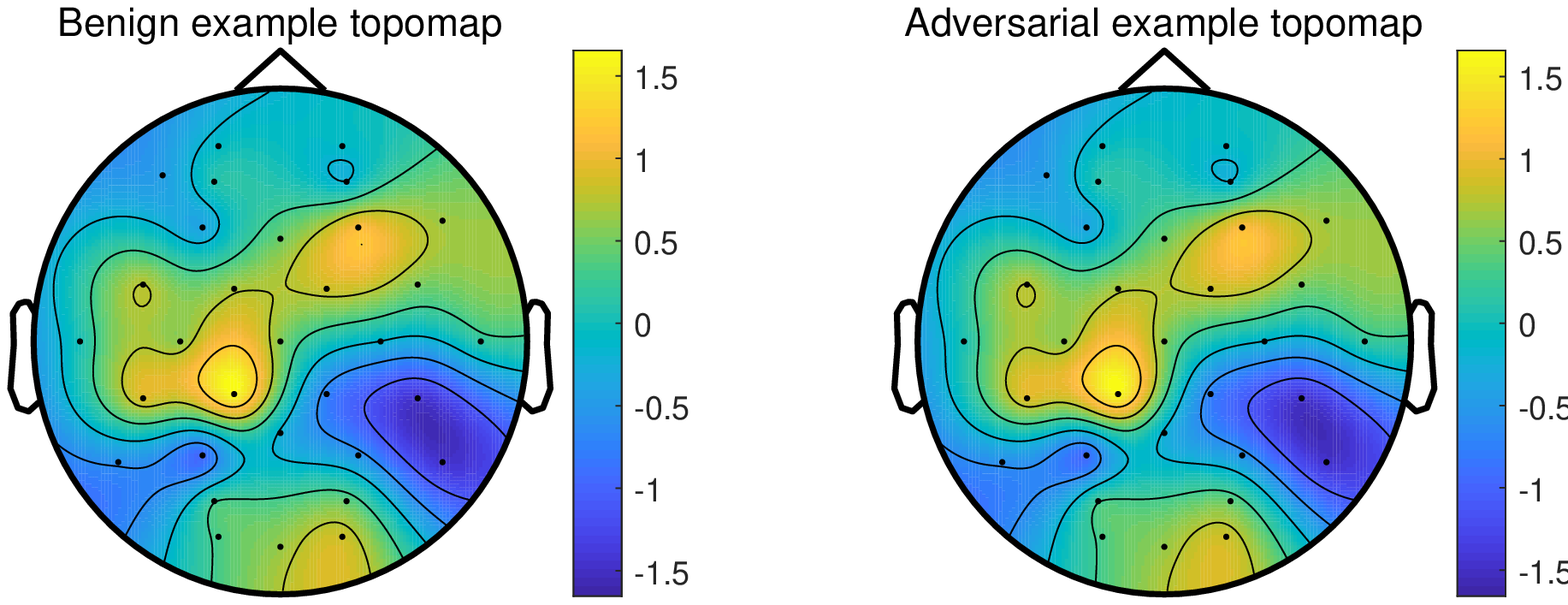}}
\caption{An example of the EEG trial before and after white-box non-target attack on the MI dataset ($\xi=0.2$). (a) time domain; (b) frequency domain; (c) topoplot.} \label{fig:examples}
\end{figure}

An example of the EEG trial before and after applying Gaussian random noise and TLM-UAP in the time domain is shown in Fig.~\ref{fig:time_example}. The signal distortion caused by the Gaussian random noise was greater than that caused by TLM-UAP under the same maximum amplitude, indicating the effectiveness of the constraint on the TLM-UAP amplitude during optimization. Fig.~\ref{fig:fre_example} shows the spectrogram of the EEG trial before and after applying TLM-UAP. The TLM-UAP was so small in both the time domain and the frequency domain that it is barely visible, and hence difficult to detect. We further show the topoplots of an EEG trial before and after white-box non-target TLM-UAP attack in Fig.~\ref{fig:topomap_example}. The difference is again very small, which may not be detectable by human eyes or a computer program.

%\begin{figure}[htpb]   \centering
%\includegraphics[width=.6\linewidth,clip]{fig3}
%\caption{An example of the EEG trial before and after the white-box non-target TLM-UAP attack on the MI dataset ($\xi=0.2$).} \label{fig:examples}
%\end{figure}

We also investigated an easily implementable channel-invariant TLM-UAP attack, which added the same perturbation to all EEG channels. The white-box attack results are shown in Table~\ref{tab:results1}. Compared with the clean and noisy baselines in Table~\ref{tab:results}, the RCAs and BCAs in Table~\ref{tab:results1} are smaller, i.e., the attacks were effective; however, the attack performances were worse than those of DF-UAP and TLM-UAP in Table~\ref{tab:results}, which is intuitive, as the channel-invariant TLM-UAP had more constraints.

\begin{table}[htbp] \center
\caption{The ratio of the number of correctly classified samples to the number of total samples (RCAs) and the average of the individual RCAs of different classes (BCAs) of the three CNN classifiers after channel-invariant TLM-UAP attacks on the three datasets ($\xi=0.2$).}  \setlength{\tabcolsep}{1.5mm}  \label{tab:results1}
\begin{indented}
\item[]\begin{tabular}{c|c|c|c|c|c|c} \toprule
\multirow{3}{*}{Experiment}&\multirow{3}{*}{Dataset}&\multirow{3}{*}{Victim Model}
&\multicolumn{4}{c}{RCA/BCA}\\ \cline{4-7}
&&&\multicolumn{2}{c|}{Baseline} &\multirow{2}{*}{TLM-UAP}  & Channel-invariant \\ \cline{4-5}
& & &Clean &Noise & &TLM-UAP \\
\midrule
&\multirow{3}{*}{P300}  &EEGNet      &$.81/.79$ &$.80/.79$ &$.17/.50$ &$.47/.66$ \\
                     &  &DeepCNN     &$.82/.78$ &$.82/.78$ &$.19/.50$ &$.60/.72$ \\
                     &  &ShallowCNN  &$.80/.75$ &$.80/.74$ &$.20/.51$ &$.64/.72$ \\ \cline{2-7}

Within &\multirow{3}{*}{ERN}&EEGNet  &$.76/.73$ &$.69/.64$ &$.30/.50$ &$.48/.58$ \\
-Subject             &  &DeepCNN     &$.69/.65$ &$.69/.65$ &$.40/.52$ &$.53/.59$ \\
                     &  &ShallowCNN  &$.70/.68$ &$.69/.67$ &$.40/.53$ &$.56/.61$ \\ \cline{2-7}

&\multirow{3}{*}{MI}    &EEGNet      &$.61/.61$ &$.47/.48$ &$.25/.25$ &$.36/.37$ \\
                     &  &DeepCNN     &$.50/.50$ &$.47/.47$ &$.26/.26$ &$.52/.51$ \\
                     &  &ShallowCNN  &$.74/.74$ &$.68/.68$ &$.25/.25$ &$.74/.74$ \\
\bottomrule
  &\multirow{3}{*}{P300}&EEGNet      &$.61/.61$ &$.47/.48$ &$.25/.25$ &$.36/.57$ \\
                     &  &DeepCNN     &$.69/.64$ &$.70/.64$ &$.18/.50$ &$.40/.58$ \\
                     &  &ShallowCNN  &$.67/.62$ &$.66/.62$ &$.19/.50$ &$.49/.61$ \\ \cline{2-7}

Cross &\multirow{3}{*}{ERN} &EEGNet  &$.67/.68$ &$.67/.65$ &$.29/.50$ &$.53/.62$ \\
-Subject             &  &DeepCNN     &$.69/.69$ &$.71/.70$ &$.31/.50$ &$.46/.56$ \\
                     &  &ShallowCNN  &$.69/.69$ &$.68/.68$ &$.29/.50$ &$.57/.61$ \\ \cline{2-7}

  &\multirow{3}{*}{MI}  &EEGNet      &$.44/.44$ &$.38/.38$ &$.25/.25$ &$.33/.33$ \\
                     &  &DeepCNN     &$.47/.47$ &$.44/.44$ &$.25/.25$ &$.33/.33$ \\
                     &  &ShallowCNN  &$.47/.47$ &$.43/.43$ &$.25/.25$ &$.38/.38$ \\
\bottomrule
\end{tabular}
\end{indented}
\end{table}

\subsection{Generalization of TLM-UAP on Traditional Classification Models}

It's also interesting to evaluate the generalization performance of the proposed TLM-UAP approach on traditional BCI classification models. We used the TLM-UAP generated from CNN models to attack some traditional non-CNN models, i.e., xDAWN~\cite{Rivet2009} spatial filtering and Logistic Regression (LR) for P300 and ERN, and common spatial pattern (CSP)~\cite{Ramoser2000} filtering and LR classifier for MI.

TLM-UAPs with three different amplitudes were generated by three CNN models, and then used to attack the traditional models. The results are shown in Table~\ref{tab:tradresult}. Observe that:
\begin{enumerate}
\item Generally, TLM-UAPs generated by CNN models were more effective to degrade the performance of the traditional models than random Gaussian noise, i.e., TLM-UAP can generalize to traditional non-CNN models.
\item The amplitude of TLM-UAP heavily affected the attack performance. Intuitively, a larger amplitude resulted in a greater model performance reduction.
\item Different models may have different resistance to TLM-UAPs. Compared with the attack performances on CNN models in Table~\ref{tab:results}, the traditional model was more robust than CNN models on the ERN dataset, but less robust on the MI dataset.
\end{enumerate}

\begin{table}[htbp] \center
\caption{The ratio of the number of correctly classified samples to the number of total samples (RCAs), and the mean RCAs of different classes (BCAs), of the traditional classifiers after TLM-UAP attacks with different $\xi$ on the three datasets.}  \setlength{\tabcolsep}{1.5mm}  \label{tab:tradresult}
\begin{indented}
\item[]\begin{tabular}{c|c|c|c|c|c|c|c} \toprule
\multirow{3}{*}{Dataset}&\multirow{3}{*}{Victim Model}
&\multirow{3}{*}{$\xi$}&\multicolumn{5}{c}{RCA/BCA}\\ \cline{4-8}
&&&\multicolumn{2}{c|}{Baseline} &\multicolumn{3}{|c}{Generation Model}\\ \cline{4-8}
& & &Clean &Noise &EEGNet &DeepCNN &ShallowCNN \\
\midrule
\multirow{3}{*}{P300}  &\multirow{3}{*}{xDAWN+LR}  &$0.05$ &$.73/.73$ &$.75/.73$ &$.66/.71$ &$.61/.70$ &$.71/.73$ \\
						&							&$0.1$ &$.73/.73$ &$.75/.73$ &$.56/.68$ &$.48/.65$ &$.66/.71$ \\
						&							&$0.2$ &$.73/.73$ &$.74/.73$ &$.38/.60$ &$.30/.56$ &$.57/.68$ \\ \midrule
\multirow{3}{*}{ERN}  &\multirow{3}{*}{xDAWN+LR}		&$0.05$ &$.67/.65$ &$.67/.65$ &$.66/.64$ &$.65/.64$ &$.66/.65$ \\
                      &                             &$0.1$ &$.67/.65$ &$.66/.65$ &$.64/.65$ &$.64/.64$ &$.65/.65$ \\
                      &                             &$0.2$ &$.67/.65$ &$.67/.65$ &$.64/.64$ &$.62/.64$ &$.63/.64$ \\ \midrule
\multirow{3}{*}{MI}  &\multirow{3}{*}{CSP+LR}		&$0.05$ &$.58/.58$ &$.57/57$ &$.47/.47$ &$.52/.52$ &$.42/.42$ \\
					& 								&$0.1$ &$.58/.58$ &$.42/.42$ &$.31/.31$ &$.32/.32$ &$.28/.28$ \\
					&								&$0.2$ &$.58/.58$ &$.27/.27$	 &$.30/.30$ &$.26/.26$ &$.27/.27$ \\

\bottomrule
\end{tabular}
\end{indented}
\end{table}

\subsection{Transferability of UAP in Gray-Box Attacks}

Transferability is one of the most threatening properties of adversarial examples, which means that adversarial examples generated by one model may also be able to attack another one. This section explores the transferability of DF-UAP and TLM-UAP.

A gray-box attack scenario was considered: the attacker only has access to the training set of the victim model, instead of its architecture and parameters. In this situation, the attacker can train a substitute model on the same training set to generate a UAP, which was then used to attack the victim model. The results are shown in the last part of Table~\ref{tab:results}. We can observe that:
\begin{enumerate}
\item The classification performances degraded after gray-box attacks, verifying the transferability of both DF-UAP and TLM-UAP.
\item In most cases, TLM-UAP led to larger classification performance degradation of the RCA and BCA than DF-UAP, suggesting that our proposed TLM-UAP was more effective than DF-UAP.
\end{enumerate}

\subsection{Characteristics of UAP} \label{sect:charUAP}

Additional experiments were performed in this subsection to analyze the characteristics of TLM-UAP.

\subsubsection{Signal-to-Perturbation Ratio (SPR)}

We computed SPRs of the perturbed EEG trials, including applying random noise, DF-UAP and TLM-UAP in white-box attacks. We treated the original EEG trials as clean signals, and computed the SPRs in cross-subject experiments. The results are shown in Table~\ref{tab:SPR}. In most cases, the SPRs of the adversarial examples perturbed by TLM-UAP were higher than those perturbed by DF-UAP, i.e., the UAP crafted by TLM had a smaller magnitude, and hence may be more difficult to detect. This is because in addition to $\|\bm{v}\|_p\leq \xi$, TLM-UAP is also bounded by the constraint function $C(\bm{x}, \bm{v})$ in (\ref{eq:expectation}).

\begin{table}[htbp] \center
\caption{SPRs (dB) of EEG trials perturbed by DF-UAP and TLM-UAP in white-box non-target attacks ($\xi=0.2$).}   \label{tab:SPR}
\begin{indented}
\item[]\begin{tabular}{c|c|ccc} \toprule
        & Dataset & EEGNet              & DeepCNN               & ShallowCNN     \\ \hline
        & P300    & $16.99$             & $17.00$               & $17.85$        \\
DF-UAP  & ERN     & $16.22$             & $16.73$               & $\mathbf{17.73}$        \\
        & MI      & $21.71$             & $13.08$               & $14.57$        \\ \hline
        & P300    & $\mathbf{21.17}$    & $\mathbf{19.92}$      & $\mathbf{20.58}$        \\
TLM-UAP & ERN     & $\mathbf{21.03}$    & $\mathbf{21.67}$      & $17.72$        \\
        & MI      & $\mathbf{23.48}$    & $\mathbf{17.85}$      & $\mathbf{17.80}$        \\
\bottomrule
\end{tabular}
\end{indented}
\end{table}

\subsubsection{Spectrogram}

\begin{figure*}[htbp]\centering
\subfigure[]{\label{fig:Spec_df}   \includegraphics[width=.9\linewidth,clip]{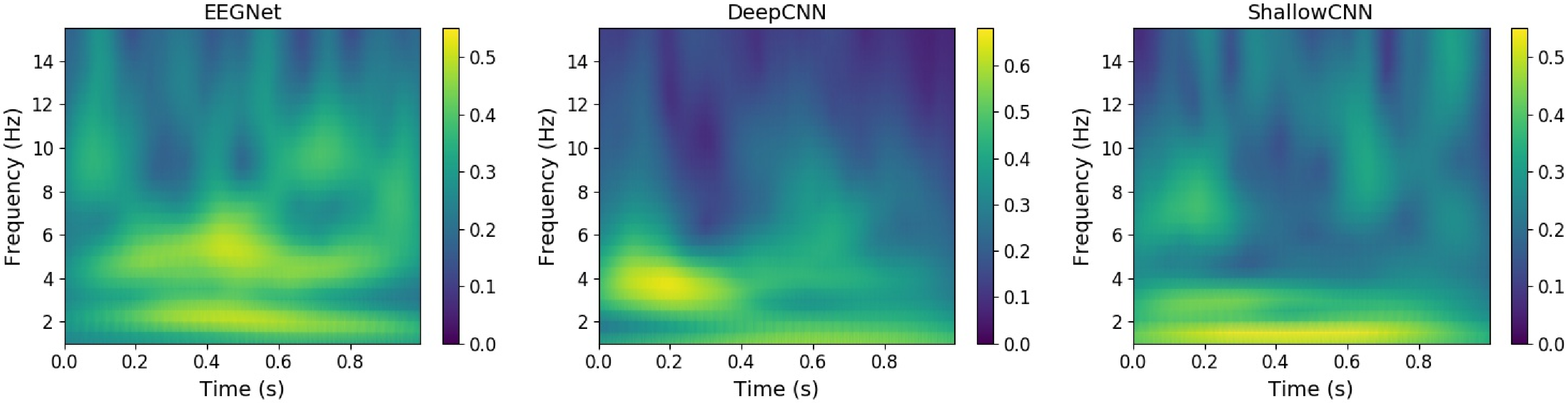}}
\subfigure[]{\label{fig:Spec_tlm}    \includegraphics[width=.92\linewidth,clip]{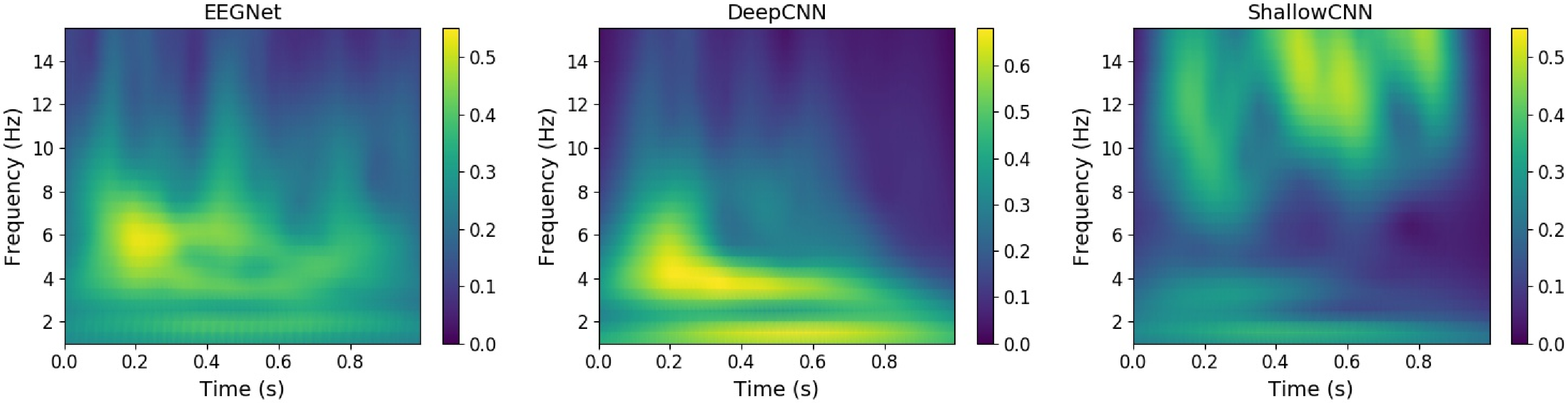}}
\caption{Spectrograms of DF-UAPs and TLM-UAPs on the P300 dataset in white-box non-target within-subject experiments. Channel \emph{$C_{z}$} was used. (a) DF-UAP; (b) TLM-UAP.} \label{fig:Spec}
\end{figure*}

In order to analyze the time-frequency characteristics of UAP, we compared the spectrograms of DF-UAP and TLM-UAP for the three classifiers in white-box attacks. The results are shown in Fig.~\ref{fig:Spec}.

DF-UAPs and TLM-UAPs share similar spectrogram patterns: for EEGNet and DeepCNN, the energy was mainly concentrated in the low-frequency areas, whereas it was more scattered for ShallowCNN. There were also some significant differences. For EEGNet, the energy of DF-UAP was concentrated in $[0.1, 0.9]$s and $[0, 7]$Hz, whereas the energy of TLM-UAP was concentrated in $[0.1, 0.8]$s and $[3, 8]$Hz. For DeepCNN, TLM-UAP seemed to affect a longer period of signals, i.e., $[0.4, 0.8]$s. For ShallowCNN, TLM-UAP mainly perturbed the high-frequency areas, which were less uniform than DF-UAP.

These results also explained the cross-subject transferability results of TLM-UAP on the P300 dataset in Table~\ref{tab:results}: TLM-UAPs generated from EEGNet and DeepCNN were more similar than those from ShallowCNN, so TLM-UAP generated from EEGNet (DeepCNN) was more effective in attacking DeepCNN (EEGNet), and their RCAs and BCAs were close.

\subsubsection{Hidden-Layer Feature Map}

\begin{figure}[htpb]   \centering
\includegraphics[width=.9\linewidth,clip]{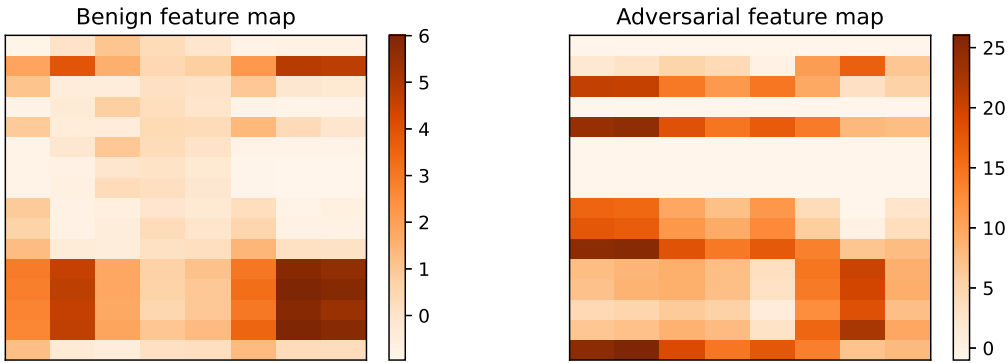}
\caption{Feature map of EEGNet before and after white-box non-target within-subject TLM-UAP attack on the MI dataset ($\xi=0.2$).} \label{fig:feature}
\end{figure}

Section~\ref{sect:White-Box-results} shows that there is no big difference between the EEG trial before and after TLM-UAP attack in the time domain and the frequency domain, and on the topoplot. Fig.~\ref{fig:feature} visualizes the feature map from the last convolution layer of EEGNet before and after adding the TLM-UAP to a clean MI EEG trial. The small perturbation was amplified by the complex nonlinear transformation of EEGNet, and hence the hidden layer feature maps were significantly different. This is intuitive, as otherwise the output of EEGNet would not change much.

\subsection{Hyper-Parameter Sensitivity}

This subsection analyzes the sensitivity of TLM-UAP to its hyper-parameters.

\subsubsection{The Magnitude of TLM-UAP}

\begin{figure*}[htpb]\centering
\subfigure[]{   \includegraphics[width=.32\columnwidth,clip]{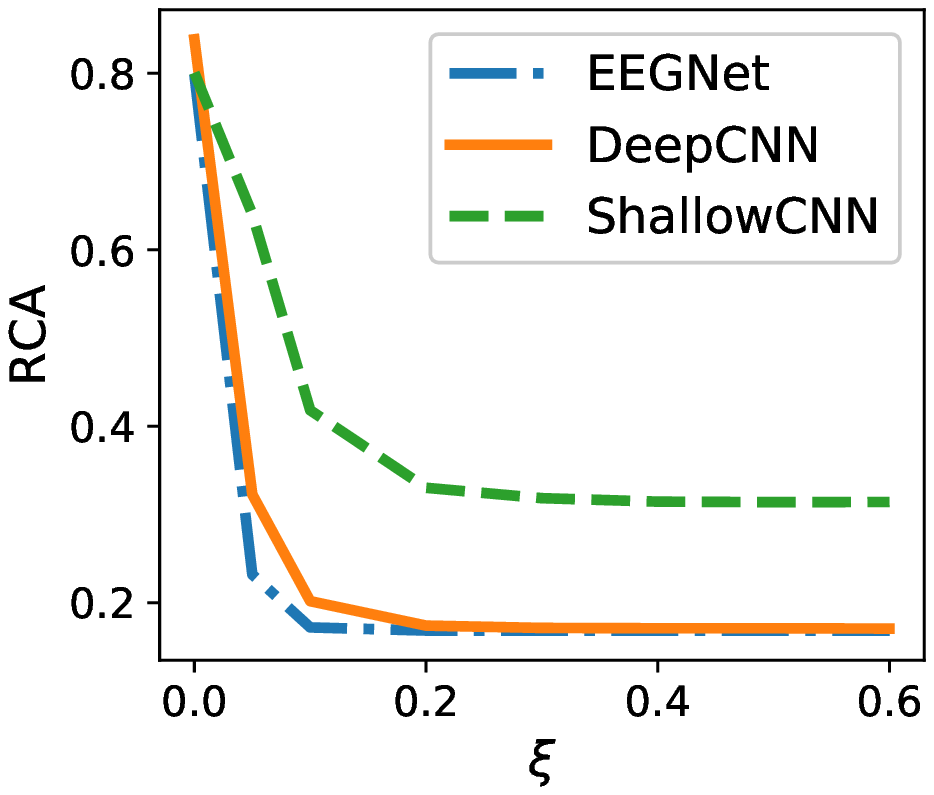}}
\subfigure[]{   \includegraphics[width=.32\columnwidth,clip]{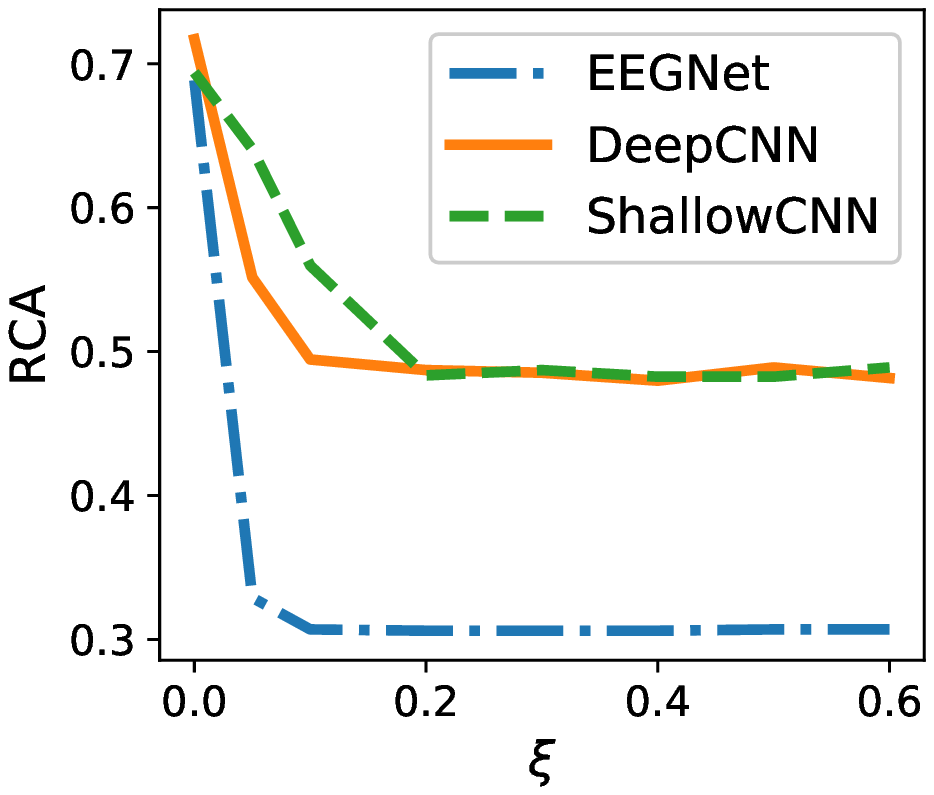}}
\subfigure[]{   \includegraphics[width=.32\columnwidth,clip]{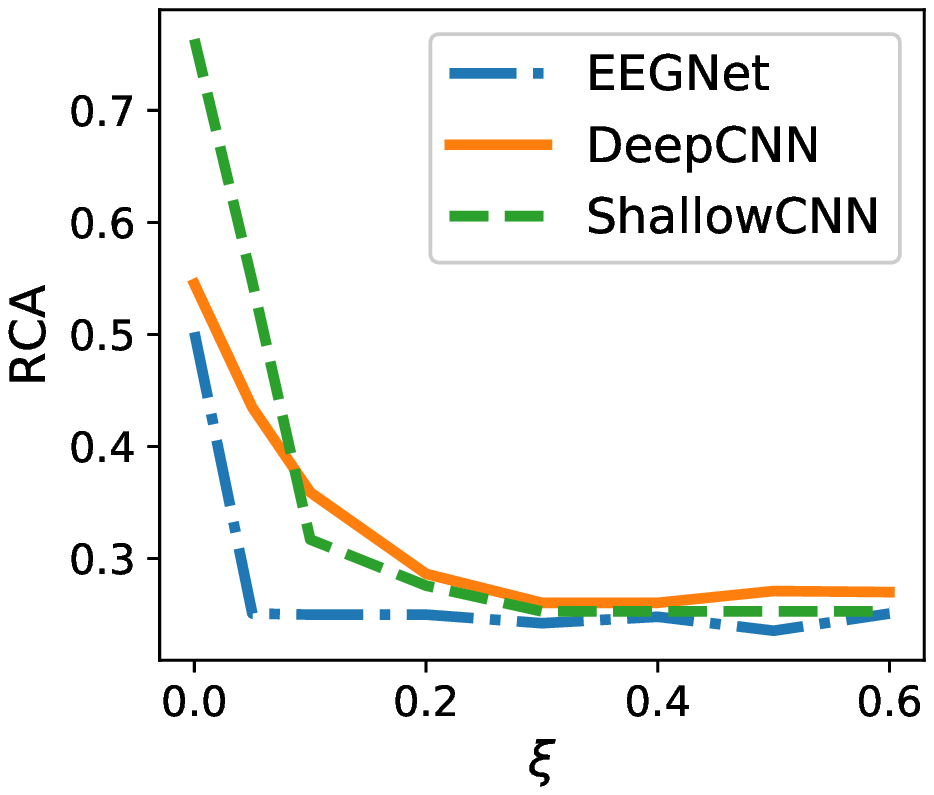}}
\caption{RCAs of the victim model after white-box non-target within-subject TLM-UAP attack, with respect to different $\xi$. (a) P300 dataset; (b) ERN dataset; and, (c) MI dataset.} \label{fig:epsilon}
\end{figure*}

$\xi$ is an important parameter in Algorithm~\ref{alg:TLM-UAP}, which directly bounds the magnitude of the perturbation. We evaluated the TLM-UAP attack performance with respect to different $\xi$. As shown in Fig.~\ref{fig:epsilon}, the RCA decreased rapidly as $\xi$ increased and converged at $\xi=0.2$ in most cases, suggesting that a small UAP is powerful enough to attack the victim model.

\subsubsection{Training Set Size}

\begin{figure}[htpb] \centering
\includegraphics[width=0.6\linewidth,clip]{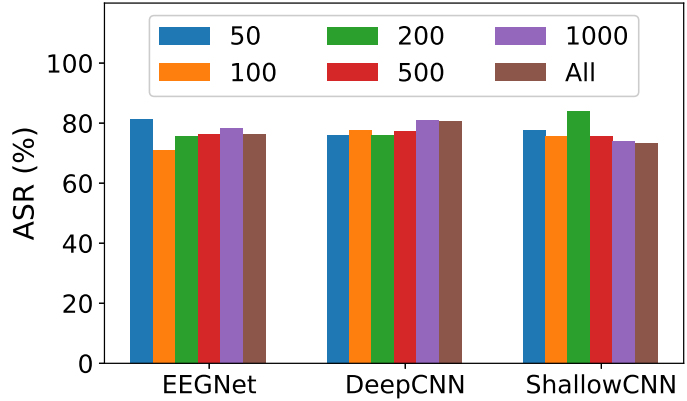}
\caption{ASRs in white-box non-target cross-subject experiment on the MI dataset, with respect to different training set sizes. `All' means all 4,608 training EEG trials in the MI dataset were used in Algorithm~\ref{alg:TLM-UAP}.} \label{fig:sample_size}
\end{figure}

It's interesting to study if the training set size affects the performance of TLM-UAP. Fig.~\ref{fig:sample_size} shows the white-box non-target attack performance of TLM-UAP, which were trained with different numbers of EEG trials in cross-subject experiments on the MI dataset. It seems that we do not need a large training set to obtain an effective TLM-UAP. The same phenomenon was also observed in \cite{Moosavi2017}.

\subsubsection{Constraint}

We also compared different constraint $C$ in (\ref{eq:expectation}): \emph{No} constraint, \emph{L1} regularization ($\alpha=10/10/5$ for EEGNet/DeepCNN/ShallowCNN), and \emph{L2} regularization ($\alpha=100$). The SPRs on the three datasets are shown in Table~\ref{tab:constraint}. Albeit similar attack performances, TLM-UAP trained with constraints led to a larger SPR (the SPRs in the `L1' and `L2' rows are larger than those in the corresponding `No' row).

\begin{table}[htbp] \centering
\caption{Mean RCAs (\%) and SPRs (dB) on the three datasets using different constraints in white-box non-target attacks ($\xi=0.2$).}   \label{tab:constraint}
\begin{indented}
\item[]\begin{tabular}{c|c|c|ccc}
\toprule
\multirow{2}{*}{Dataset} & \multirow{2}{*}{Constraint} & \multirow{2}{*}{Mean RCA} & \multicolumn{3}{c}{SPR of TLM-UAP}   \\ \cline{4-6}
                         &               &                    & EEGNet           & DeepCNN          & ShallowCNN          \\ \hline
\multirow{3}{*}{P300}    & No             & $17.18$            & $14.89$          & $14.71$          & $14.45$             \\
                         & L1            & $17.36$            & $18.39$          & $17.82$          & $17.16$             \\
                         & L2            & $17.85$            & $21.17$          & $19.92$          & $20.58$             \\ \hline
\multirow{3}{*}{ERN}     & No             & $30.96$            & $19.91$          & $20.70$          & $17.02$             \\
                         & L1            & $29.24$            & $21.45$          & $22.05$          & $17.11$             \\
                         & L2            & $30.66$            & $21.03$          & $21.67$          & $17.72$             \\ \hline
\multirow{3}{*}{MI}      & No             & $25.05$            & $22.88$          & $15.46$          & $16.11$             \\
                         & L1            & $25.08$            & $23.35$          & $53.76$          & $16.88$             \\
                         & L2            & $25.06$            & $23.48$          & $17.85$          & $17.80$             \\

\bottomrule
\end{tabular}
\end{indented}
\end{table}

\begin{figure}[htpb]\centering
\includegraphics[width=0.6\linewidth,clip]{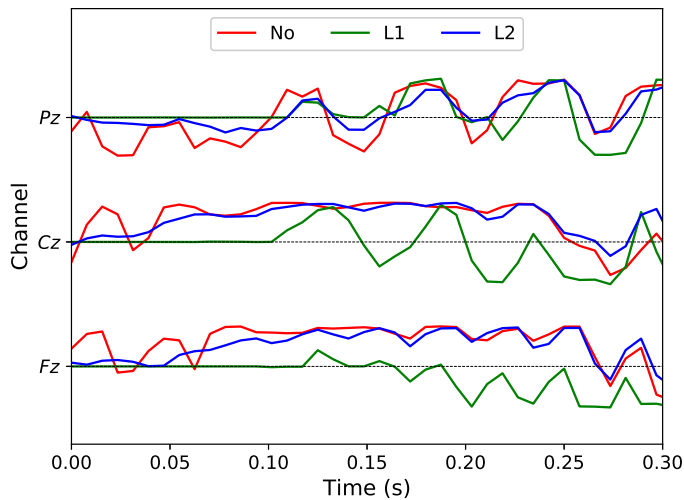}
\caption{TLM-UAP trained with different constraints on the MI dataset in white-box non-target attacks. Channels \emph{$P_{z}$}, \emph{$C_{z}$} and \emph{$F_{z}$} were used.} \label{fig:reg}
\end{figure}

Fig.~\ref{fig:reg} shows that adding different constraints significantly changed the waveforms of TLM-UAP. \emph{L1} regularization introduced sparsity, whereas \emph{L2} regularization reduced the perturbation magnitude.

We may also generate a TLM-UAP which satisfies other requirements by changing the constraint function $C$, such as perturbing certain EEG channels, or even against a metric function which is used to detect adversarial examples. We will leave these to our future research.

\subsection{Influence of the Batch Size}

Mini-batch gradient descent was used to optimize TLM-UAP. Table~\ref{tab:batch_size_result} shows the change of RCAs and BCAs w.r.t. different batch sizes. Generally, the proposed TLM-UAP is insensitive to batch size within $[16, 64]$.

\begin{table}[htbp] \center
\caption{The ratio of the number of correctly classified samples to the number of total samples (RCAs), and the mean RCAs of different classes (BCAs), of the three CNN classifiers after TLM-UAP attacks with different batch sizes on the three datasets.}  \setlength{\tabcolsep}{1.5mm}  \label{tab:batch_size_result}
\begin{indented}
\item[]\begin{tabular}{c|c|c|c|c} \toprule
\multirow{3}{*}{Dataset}&\multirow{3}{*}{Victim Model}
&\multicolumn{3}{c}{RCA/BCA}\\ \cline{3-5}
& &Batch Size $16$ &Batch Size $32$ &Batch Size $64$ \\
\midrule
\multirow{3}{*}{P300}  &EEGNet  	&$.17/.50$ &$.17/.50$ &$.17/.50$ \\
					   &DeepCNN 	&$.17/.50$ &$.17/.50$ &$.19/.50$ \\
					   &ShallowCNN  &$.21/.51$ &$.20/.51$ &$.20/.51$ \\ \midrule
\multirow{3}{*}{ERN}   &EEGNet  	&$.31/.50$ &$.31/.50$ &$.30/.50$ \\
					   &DeepCNN 	&$.41/.52$ &$.40/.52$ &$.40/.52$ \\
					   &ShallowCNN  &$.38/.51$ &$.38/.51$ &$.40/.53$ \\ \midrule
\multirow{3}{*}{MI}  &EEGNet  	&$.25/.25$ &$.25/.25$ &$.25/.25$ \\
					   &DeepCNN 	&$.29/.29$ &$.29/.29$ &$.26/.26$ \\
					   &ShallowCNN  &$.26/.26$ &$.26/.26$ &$.25/.25$ \\ \bottomrule
\end{tabular}
\end{indented}
\end{table}

\section{Target Attack Results} \label{sect:results2}

Our TLM approach is also capable of performing target attacks, which can be easily achieved by changing the loss function $l$ in (\ref{eq:expectation}).

We performed white-box target attacks in cross-subject experiments on the three datasets and evaluated the \emph{target rate}, which is the number of samples classified to the target class divided by the number of total samples. The results are shown in Table~\ref{tab:target_results}. TLM-UAPs had close to 100\% target rates in white-box target attacks, indicating that our approach can manipulate the BCI systems to output whatever the attacker wants, which may be more dangerous than non-target attacks. For example, in a BCI-driven wheelchair, a target TLM-UAP attack may force all commands to be interpreted as a specific command (e.g., going forward), and hence run the user into danger.

\begin{table}[htbp] \center
\caption{Target rates of the three CNN classifiers in cross-subject white-box target TLM-UAP attacks on the three datasets ($\xi=0.2$).}  \setlength{\tabcolsep}{1.3mm}  \label{tab:target_results}
\begin{indented}
\item[]\begin{tabular}{c|c|c|c|c|c} \toprule
\multirow{2}{*}{Dataset} & \multirow{2}{*}{Victim Model} & \multirow{2}{*}{Target Class} & \multicolumn{2}{c|}{Baseline} & TLM-UAP \\ \cline{4-5}
                         &      &                        & Clean         & Noisy         & Attack \\ \hline
                        \multirow{6}{*}{P300}       & \multirow{2}{*}{EEGNet}       & Non-target                        & $.6627$         & $.6463$        & $.9629$ \\
                                                    &                               & Target                            & $.3373$         & $.3510$        & $.9572$ \\ \cline{2-6}
                                                    & \multirow{2}{*}{DeepCNN}      & Non-target                        & $.6755$         & $.6637$        & $.9416$ \\
                                                    &                               & Target                            & $.3245$         & $.3116$        & $.9373$ \\ \cline{2-6}
                                                    & \multirow{2}{*}{ShallowCNN}   & Non-target                        & $.6505$         & $.6597$        & $.8904$ \\
                                                    &                               & Target                            & $.3495$         & $.3499$        & $.8306$ \\ \cline{1-6}
                        \multirow{6}{*}{ERN}        & \multirow{2}{*}{EEGNet}       & Bad                               & $.3537$         & $.3741$        & $.9980$ \\
                                                    &                               & Good                              & $.6463$         & $.6300$        & $.9971$ \\ \cline{2-6}
                                                    & \multirow{2}{*}{DeepCNN}      & Bad                               & $.3770$         & $.3309$        & $.9912$ \\
                                                    &                               & Good                              & $.6230$         & $.6739$        & $.9976$ \\ \cline{2-6}
                                                    & \multirow{2}{*}{ShallowCNN}   & Bad                               & $.3033$         & $.2910$        & $.9741$ \\
                                                    &                               & Good                              & $.6967$         & $.7160$        & $.9888$ \\ \cline{1-6}
                        \multirow{12}{*}{MI}        & \multirow{4}{*}{EEGNet}       & Left                              & $.3152$         & $.1350$        & $.9821$ \\
                                                    &                               & Right                             & $.2830$         & $.2056$        & $.9850$ \\
                                                    &                               & Feet                              & $.1545$         & $.1954$        & $.9994$ \\
                                                    &                               & Tongue                            & $.2473$         & $.5380$        & $1.000$ \\ \cline{2-6}
                                                    & \multirow{4}{*}{DeepCNN}      & Left                              & $.2535$         & $.1765$        & $.8839$ \\
                                                    &                               & Right                             & $.3491$         & $.2207$        & $.9238$ \\
                                                    &                               & Feet                              & $.2282$         & $.3155$        & $.9659$ \\
                                                    &                               & Tongue                            & $.1692$         & $.2544$        & $.9938$ \\ \cline{2-6}
                                                    & \multirow{4}{*}{ShallowCNN}   & Left                              & $.2872$         & $.1952$        & $.9151$ \\
                                                    &                               & Right                             & $.2537$         & $.1746$        & $.9443$ \\
                                                    &                               & Feet                              & $.2647$         & $.2838$        & $.9819$ \\
                                                    &                               & Tongue                            & $.2124$         & $.3673$        & $.9983$ \\
\bottomrule
\end{tabular}
\end{indented}
\end{table}

%To our knowledge, no one has studied optimization based UAPs for target attacks before. Our TLM approach is the first in this direction.

\begin{figure*}[htpb]\centering
\subfigure[]{   \includegraphics[width=.32\columnwidth,clip]{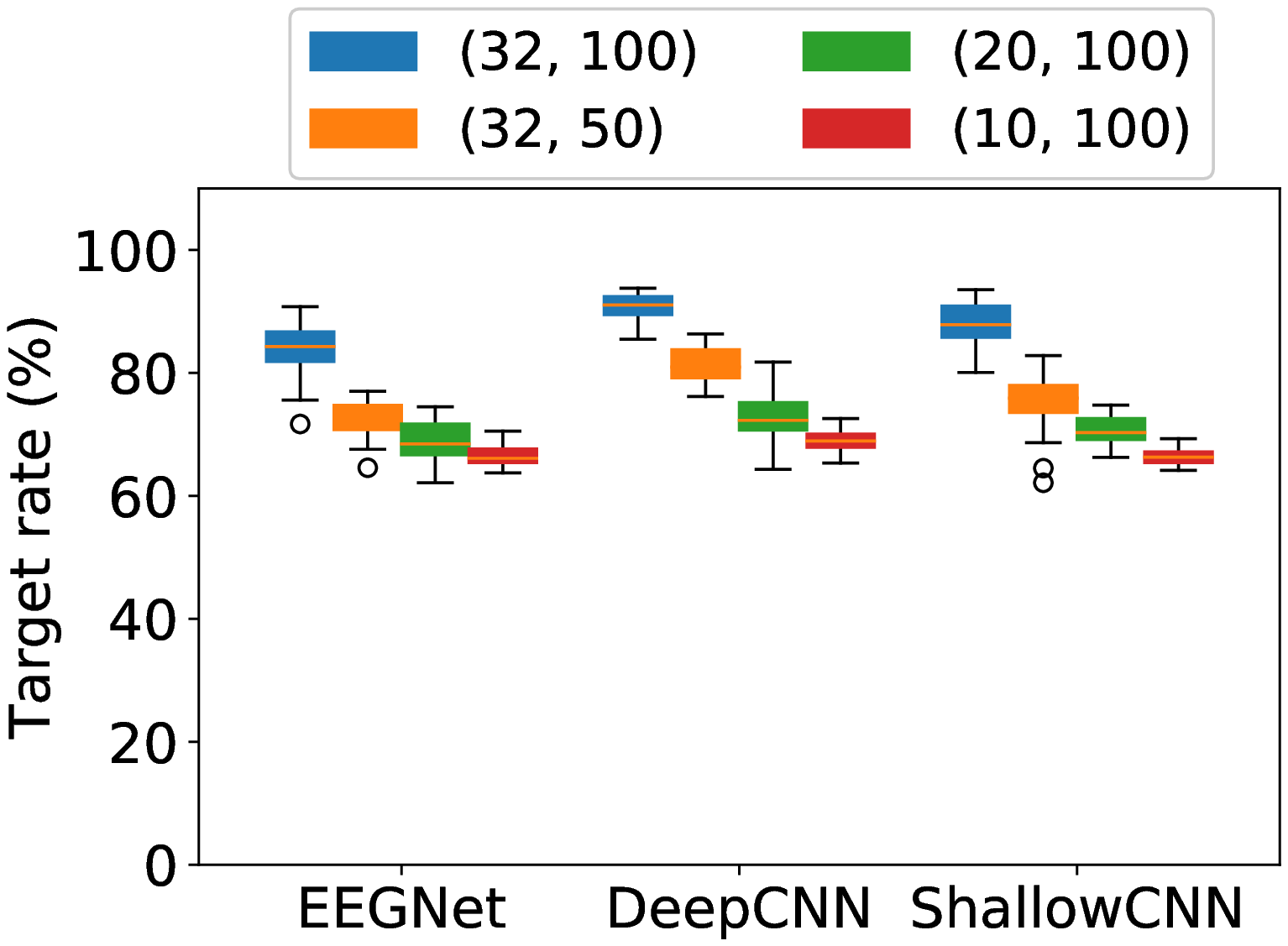}}
\subfigure[]{   \includegraphics[width=.32\columnwidth,clip]{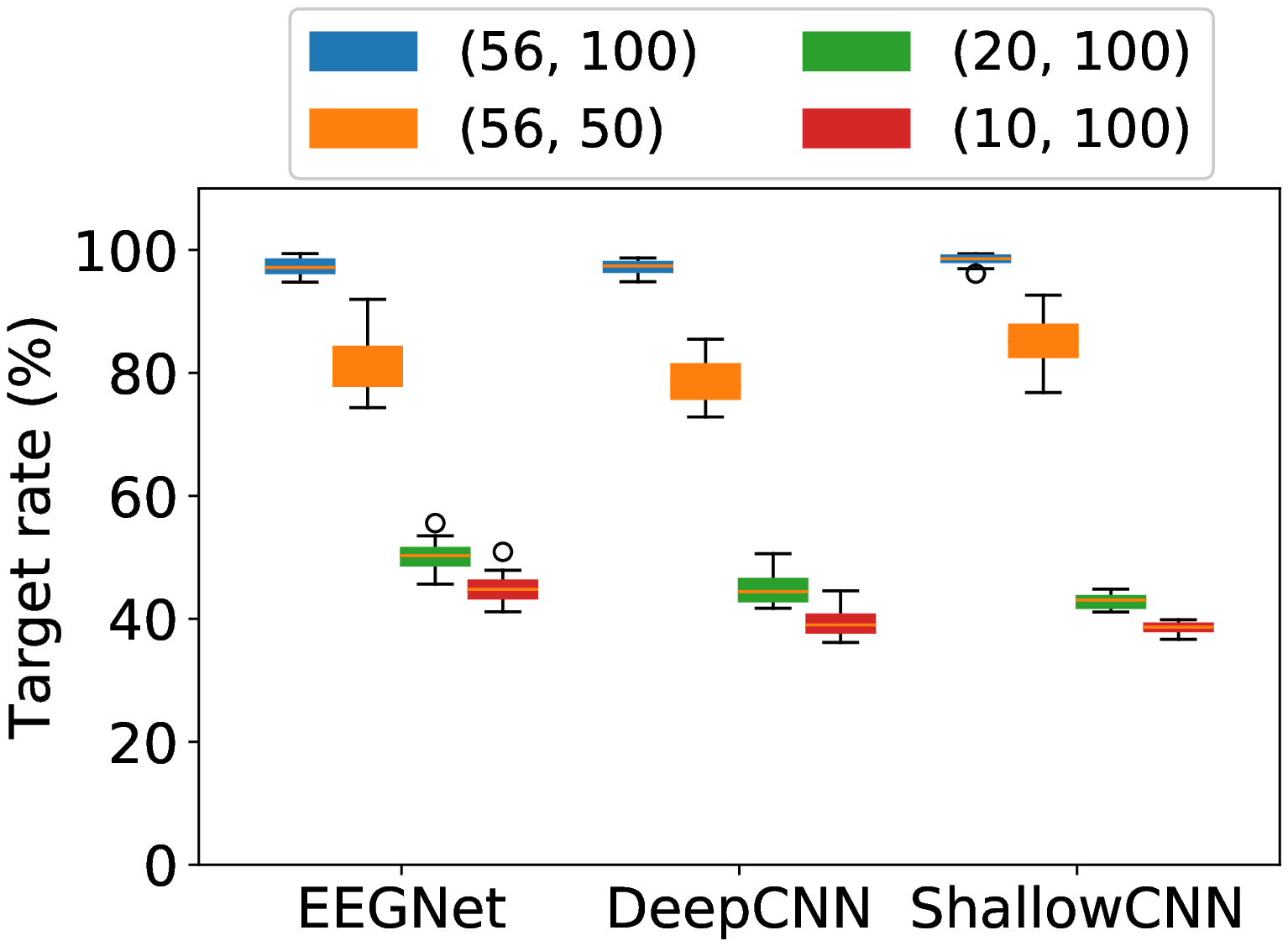}}
\subfigure[]{   \includegraphics[width=.32\columnwidth,clip]{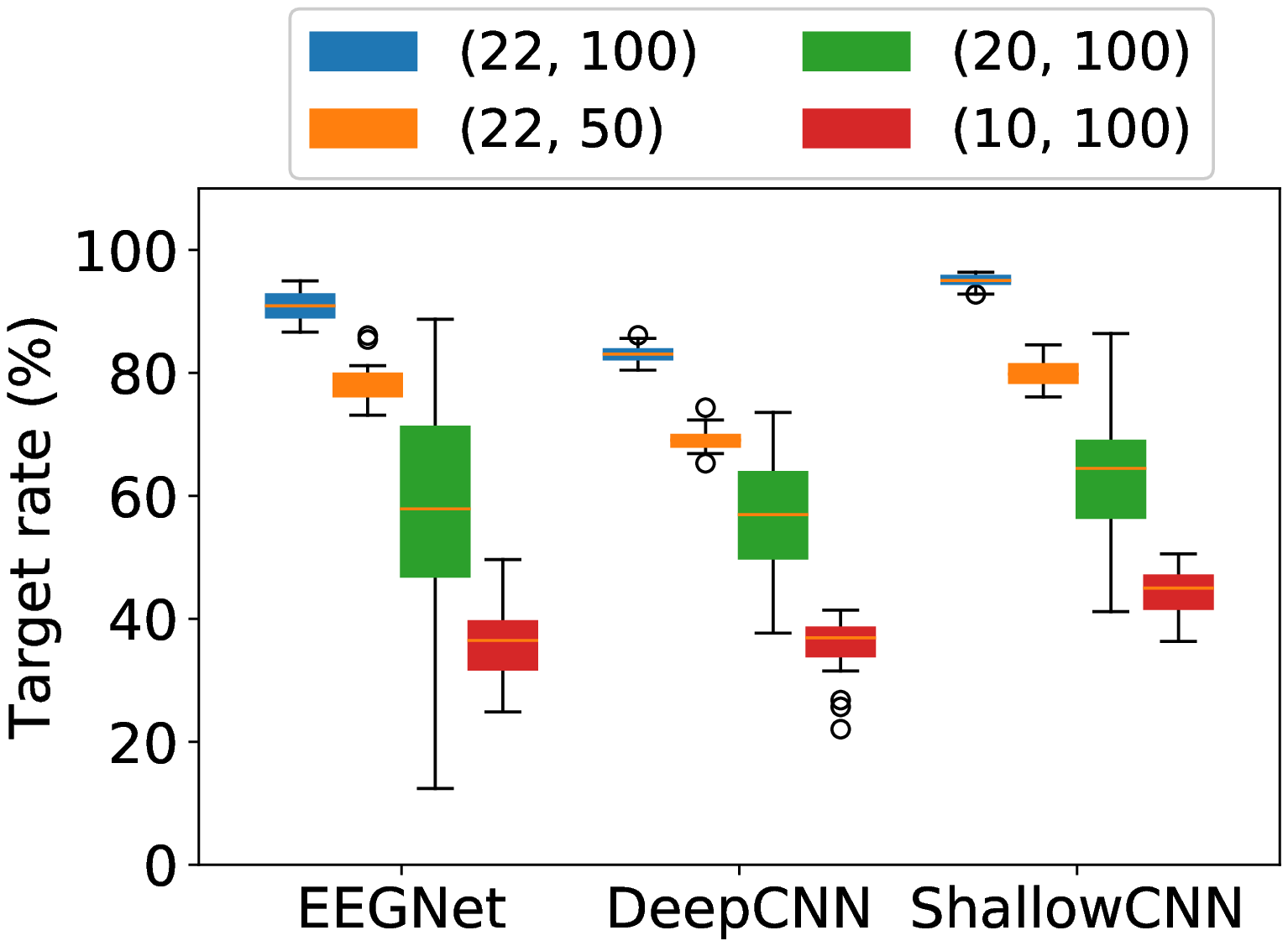}}
\caption{Target rates of cross-subject mini TLM-UAP target attacks, with different UAP template size $(C_m, T_m)$, where $C_m$ is the number of EEG channels and $T_m$ the number of time domain samples. The original trial sizes of P300/ERN/MI datasets were $(32, 256)/(56, 260)/(22, 256)$, respectively. (a) P300 dataset, non-target class; (b) ERN dataset, bad-feedback class; (c) MI dataset, left-hand class.} \label{fig:patch_size}
\end{figure*}

To further simplify the implementation of TLM-UAP, we also considered smaller template size, i.e., mini TLM-UAP with a small number of channels and time domain samples, which can be added anywhere to an EEG trail. Mini TLM-UAPs are more practical and flexible, because they do not require the attacker to know the exact number of EEG channels and the exact length and starting time of an EEG trial. During optimization, we randomly placed the mini TLM-UAP at different locations (both channel-wise and time-wise) of EEG trials and tried to make the attacks successful. During test, the mini TLM-UAP was randomly added to 30 different locations of each EEG trail. The results are shown in Fig.~\ref{fig:patch_size}. Generally, all mini TLM-UAPs were effective. However, their effectiveness decreased when the number of used channels ($C_m$) and/or the template length ($T_m$) decreased, which is intuitive. These results suggest that a mini TLM-UAP may be used to achieve a better compromise between the attack performance and the implementation difficulty.

\section{Conclusions and Future Research} \label{sect:conclusions}

Multiple CNN classifiers have been proposed for EEG-based BCIs. However, CNN models are vulnerable to UAPs, which are small and example-independent perturbations, yet powerful enough to significantly degrade the performance of a CNN model when added to a benign example. This paper has proposed a novel TLM approach to generate UAP for EEG-based BCI systems. Experimental results demonstrated its effectiveness in attacking three popular CNN classifiers for both non-target and target attacks. We also verified the transferability of the UAPs in EEG-based BCI systems. To our knowledge, this is the first study on UAPs of CNN classifiers in EEG-based BCIs. It exposes a potentially critical security problem in BCIs, and hopefully will lead to the design of safer BCIs.

Our future research will enhance the transferability of TLM-UAP in deep learning, and also consider how to attack traditional machine learning models in EEG-based BCIs. More importantly, we will design effective strategies to defend against UAP attacks. Multiple approaches, e.g., adversarial training~\cite{FGSM}, defensive distillation~\cite{AdvTransferPapernot2016}, ensemble adversarial training~\cite{EnsembleAdvTraining}, and so on~\cite{Madry2018, Xie, AdvInputTransform,Akhtar2018}, have been proposed to defend against adversarial examples in other application domains. As TLM is a perturbation based first-order gradient optimization approach, PGD~\cite{Madry2018} training may be used to defend against it.

\section*{References}
%\bibliographystyle{unsrt}\bibliography{zhliu}

\begin{thebibliography}{10}

\bibitem{P300Sutton}
Samuel Sutton, Margery Braren, Joseph Zubin, and E.~R. John.
\newblock Evoked-potential correlates of stimulus uncertainty.
\newblock {\em Science}, 150(3700):1187--1188, 1965.

\bibitem{P3001988}
LA~Farwell and E~Donchin.
\newblock Talking off the top of your head: {T}oward a mental prosthesis
  utilizing event-related brain potentials.
\newblock {\em Electroencephalography and Clinical Neurophysiology},
  70(6):510--523, December 1988.

\bibitem{drwuTHMS2017}
Dongrui Wu.
\newblock Online and offline domain adaptation for reducing {BCI} calibration
  effort.
\newblock {\em {IEEE} Trans. on Human-Machine Systems}, 47(4):550--563, 2017.

\bibitem{drwuTNSRE2016}
Dongrui Wu, Vernon~J. Lawhern, W.~D. Hairston, and Brent~J. Lance.
\newblock Switching {EEG} headsets made easy: {Reducing} offline calibration
  effort using active weighted adaptation regularization.
\newblock {\em {IEEE} Trans. on Neural Systems and Rehabilitation Engineering},
  24(11):1125--1137, 2016.

\bibitem{MI2001}
G.~Pfurtscheller and C.~Neuper.
\newblock Motor imagery and direct brain-computer communication.
\newblock {\em Proc. {IEEE}}, 89(7):1123--1134, Jul 2001.

\bibitem{SSVEPSurvey}
Danhua Zhu, Jordi Bieger, Gary Garcia~Molina, and Ronald~M Aarts.
\newblock A survey of stimulation methods used in {SSVEP}-based {BCI}s.
\newblock {\em Computational Intelligence and Neuroscience}, page 702357, 2010.

\bibitem{EEGNet}
Vernon~J. Lawhern, Amelia~J. Solon, Nicholas~R. Waytowich, Stephen~M. Gordon,
  Chou~P. Hung, and Brent~J. Lance.
\newblock {EEGNet}: {A} compact convolutional neural network for {EEG}-based
  brain-computer interfaces.
\newblock {\em Journal of Neural Engineering}, 15(5):056013, June 2018.

\bibitem{MNE}
Robin~Tibor Schirrmeister, Jost~Tobias Springenberg, Lukas Dominique~Josef
  Fiederer, Martin Glasstetter, Katharina Eggensperger, Michael Tangermann,
  Frank Hutter, Wolfram Burgard, and Tonio Ball.
\newblock Deep learning with convolutional neural networks for {EEG} decoding
  and visualization.
\newblock {\em Human Brain Mapping}, 38(11):5391--5420, 2017.

\bibitem{EEG2Image}
Pouya Bashivan, Irina Rish, Mohammed Yeasin, and Noel Codella.
\newblock Learning representations from {EEG} with deep recurrent-convolutional
  neural networks.
\newblock In {\em Proc. Int'l Conf. on Learning Representations}, San Juan,
  Puerto Rico, May 2016.

\bibitem{DeepLearningMI}
Yousef~Rezaei Tabar and Ugur Halici.
\newblock A novel deep learning approach for classification of {EEG} motor
  imagery signals.
\newblock {\em Journal of Neural Engineering}, 14(1):016003, 2017.

\bibitem{Tayeb2019}
Zied Tayeb, Juri Fedjaev, Nejla Ghaboosi, Christoph Richter, Lukas Everding,
  Xingwei Qu, Yingyu Wu, Gordon Cheng, and J{\"o}rg Conradt.
\newblock Validating deep neural networks for online decoding of motor imagery
  movements from {EEG} signals.
\newblock {\em Sensors}, 19(1):210, January 2019.

\bibitem{AdvExamSzegedy}
Christian Szegedy, Wojciech Zaremba, Ilya Sutskever, Joan Bruna, Dumitru Erhan,
  Ian~J. Goodfellow, and Rob Fergus.
\newblock Intriguing properties of neural networks.
\newblock In {\em Proc. Int'l Conf. on Learning Representations}, Banff,
  Canada, April 2014.

\bibitem{Biggio2013}
Battista Biggio, Igino Corona, Davide Maiorca, Blaine Nelson, Nedim
  {\v{S}}rndi{\'c}, Pavel Laskov, Giorgio Giacinto, and Fabio Roli.
\newblock Evasion attacks against machine learning at test time.
\newblock In {\em Proc. Joint European Conf. on Machine Learning and Knowledge
  Discovery in Databases}, pages 387--402, Berlin, Germany, September 2013.

\bibitem{FGSM}
Ian~J. Goodfellow, Jonathon Shlens, and Christian Szegedy.
\newblock Explaining and harnessing adversarial examples.
\newblock In {\em Proc. Int'l Conf. on Learning Representations}, San Diego,
  CA, May 2015.

\bibitem{BIM}
Alexey Kurakin, Ian~J. Goodfellow, and Samy Bengio.
\newblock Adversarial examples in the physical world.
\newblock In {\em Proc. Int'l Conf. on Learning Representations}, Toulon,
  France, April 2017.

\bibitem{AdvPatch}
Tom~B. Brown, Dandelion Man{\'{e}}, Aurko Roy, Mart{\'{\i}}n Abadi, and Justin
  Gilmer.
\newblock Adversarial patch.
\newblock {\em CoRR}, abs/1712.09665, 2017.

\bibitem{Adv3D}
Anish Athalye, Logan Engstrom, Andrew Ilyas, and Kevin Kwok.
\newblock Synthesizing robust adversarial examples.
\newblock In {\em Proc. 35th Int'l Conf. on Machine Learning}, pages 284--293,
  Stockholm, Sweden, July 2018.

\bibitem{Carlini2018}
Nicholas Carlini and David~A. Wagner.
\newblock Audio adversarial examples: Targeted attacks on speech-to-text.
\newblock In {\em Proc. {IEEE} Symposium on Security and Privacy}, pages 1--7,
  San Francisco, CA, May 2018.

\bibitem{Grosse2016}
Kathrin Grosse, Nicolas Papernot, Praveen Manoharan, Michael Backes, and
  Patrick McDaniel.
\newblock Adversarial perturbations against deep neural networks for malware
  classification.
\newblock {\em CoRR}, abs/1606.04435, 2016.

\bibitem{Moosavi2016}
Seyed-Mohsen Moosavi-Dezfooli, Alhussein Fawzi, and Pascal Frossard.
\newblock Deepfool: {A} simple and accurate method to fool deep neural
  networks.
\newblock In {\em Proc. {IEEE} Conf. on Computer Vision and Pattern
  Recognition}, pages 2574--2582, Las Vegas, NV, June 2016.

\bibitem{Moosavi2017}
Seyed-Mohsen Moosavi-Dezfooli, Alhussein Fawzi, Omar Fawzi, and Pascal
  Frossard.
\newblock Universal adversarial perturbations.
\newblock In {\em Proc. {IEEE} Conf. on Computer Vision and Pattern
  Recognition}, pages 1765--1773, Honolulu, HI, July 2017.

\bibitem{Baluja2017}
Shumeet Baluja and Ian Fischer.
\newblock Adversarial transformation networks: Learning to generate adversarial
  examples.
\newblock {\em CoRR}, abs/1703.09387, 2017.

\bibitem{Madry2018}
Aleksander Madry, Aleksandar Makelov, Ludwig Schmidt, Dimitris Tsipras, and
  Adrian Vladu.
\newblock Towards deep learning models resistant to adversarial attacks.
\newblock In {\em Proc. Int'l. Conf. on Learning Representations}, Vancouver,
  Canada, May 2018.

\bibitem{AdvCW}
Nicholas Carlini and David Wagner.
\newblock Towards evaluating the robustness of neural networks.
\newblock In {\em Proc. {IEEE} Symposium on Security and Privacy}, pages
  39--57, San Jose, CA, May 2017. IEEE.

\bibitem{Zhang2019}
X.~{Zhang} and D.~{Wu}.
\newblock On the vulnerability of {CNN} classifiers in {EEG}-based {BCI}s.
\newblock {\em {IEEE} Trans. on Neural Systems and Rehabilitation Engineering},
  27(5):814--825, May 2019.

\bibitem{Li2016}
Jundong Li, Kewei Cheng, Suhang Wang, Fred Morstatter, Robert~P. Trevino,
  Jiliang Tang, and Huan Liu.
\newblock Feature selection: {A} data perspective.
\newblock {\em CoRR}, abs/1601.07996, 2016.

\bibitem{drwuNSR2021}
Xiao Zhang, Dongrui Wu, Lieyun Ding, Hanbin Luo, Chin-Teng Lin, Tzyy-Ping Jung,
  and Ricardo Chavarriaga.
\newblock Tiny noise, big mistakes: Adversarial perturbations induce errors in
  brain-computer interface spellers.
\newblock {\em National Science Review}, 8(4), 2021.

\bibitem{Neekhara2019}
Paarth Neekhara, Shehzeen Hussain, Prakhar Pandey, Shlomo Dubnov, Julian~J.
  McAuley, and Farinaz Koushanfar.
\newblock Universal adversarial perturbations for speech recognition systems.
\newblock {\em CoRR}, abs/1905.03828, 2019.

\bibitem{Behjati2019}
Melika Behjati, Seyed-Mohsen Moosavi-Dezfooli, Mahdieh~Soleymani Baghshah, and
  Pascal Frossard.
\newblock Universal adversarial attacks on text classifiers.
\newblock In {\em Proc. {IEEE} Int'l Conf. on Acoustics, Speech and Signal
  Processing}, pages 7345--7349, Brighton, United Kingdom, May 2019.

\bibitem{Mopuri2019}
Konda~Reddy Mopuri, Aditya Ganeshan, and Venkatesh~Babu Radhakrishnan.
\newblock Generalizable data-free objective for crafting universal adversarial
  perturbations.
\newblock {\em {IEEE} Trans. on Pattern Analysis and Machine Intelligence},
  41(10):2452--2465, October 2019.

\bibitem{Hirano2020}
Hokuto Hirano and Kazuhiro Takemoto.
\newblock Simple iterative method for generating targeted universal adversarial
  perturbations.
\newblock In {\em Proc. of 25th Int'l. Symposium on Artificial Life and
  Robotics}, pages 426--430, Beppu, Japan, January 2020.

\bibitem{drwuALBCI2019}
Xue Jiang, Xiao Zhang, and Dongrui Wu.
\newblock Active learning for black-box adversarial attacks in {EEG}-based
  brain-computer interfaces.
\newblock In {\em Proc. IEEE Symposium Series on Computational Intelligence},
  Xiamen, China, December 2019.

\bibitem{EPFLP300}
Ulrich Hoffmann, Jean-Marc Vesin, Touradj Ebrahimi, and Karin Diserens.
\newblock An efficient {P}300-based brain-computer interface for disabled
  subjects.
\newblock {\em Journal of Neuroscience Methods}, 167(1):115--125, January 2008.

\bibitem{ERN}
Perrin Margaux, Maby Emmanuel, Daligault Sébastien, Bertrand Olivier, and
  Mattout Jérémie.
\newblock Objective and subjective evaluation of online error correction during
  {P}300-based spelling.
\newblock {\em Advances in Human-Computer Interaction}, 2012(578295):13,
  October 2012.

\bibitem{MI4C}
Michael Tangermann, Klaus-Robert Müller, Ad~Aertsen, Niels Birbaumer,
  Christoph Braun, Clemens Brunner, Robert Leeb, Carsten Mehring, Kai Miller,
  Gernot Mueller-Putz, Guido Nolte, Gert Pfurtscheller, Hubert Preissl, Gerwin
  Schalk, Alois Schlögl, Carmen Vidaurre, Stephan Waldert, and Benjamin
  Blankertz.
\newblock Review of the {BCI} {C}ompetition {IV}.
\newblock {\em Frontiers in Neuroscience}, 6:55, 2012.

\bibitem{Xception}
Francois Chollet.
\newblock {X}ception: Deep learning with depthwise separable convolutions.
\newblock In {\em Proc. IEEE Conf. on Computer Vision and Pattern Recognition},
  pages 1800--1807, Honolulu, HI, July 2017.

\bibitem{FBCSP}
Kai~Keng Ang, Zheng~Yang Chin, Haihong Zhang, and Cuntai Guan.
\newblock Filter bank common spatial pattern ({FBCSP}) in brain-computer
  interface.
\newblock In {\em Proc. {IEEE} Int'l Joint Conf. on Neural Networks}, Hong
  Kong, China, June 2008.

\bibitem{Adam}
Diederik~P. Kingma and Jimmy Ba.
\newblock Adam: {A} method for stochastic optimization.
\newblock {\em CoRR}, abs/1412.6980, 2014.

\bibitem{Mann1947}
Henry~B Mann and Donald~R Whitney.
\newblock On a test of whether one of two random variables is stochastically
  larger than the other.
\newblock {\em The annals of mathematical statistics}, pages 50--60, 1947.

\bibitem{Rivet2009}
B.~Rivet, A.~Souloumiac, V.~Attina, and G.~Gibert.
\newblock {xDAWN} algorithm to enhance evoked potentials: application to
  brain-computer interface.
\newblock {\em {IEEE} Trans. on Biomedical Engineering}, 56(8):2035--2043,
  2009.

\bibitem{Ramoser2000}
Herbert Ramoser, Johannes Muller-Gerking, and Gert Pfurtscheller.
\newblock Optimal spatial filtering of single trial {EEG} during imagined hand
  movement.
\newblock {\em {IEEE} Trans. on Rehabilitation Engineering}, 8(4):441--446,
  2000.

\bibitem{AdvTransferPapernot2016}
Nicolas Papernot, Patrick McDaniel, Xi~Wu, Somesh Jha, and Ananthram Swami.
\newblock Distillation as a defense to adversarial perturbations against deep
  neural networks.
\newblock In {\em Proc. {IEEE} Symposium on Security and Privacy}, pages
  582--597, San Jose, CA, May 2016. IEEE.

\bibitem{EnsembleAdvTraining}
Florian Tramèr, Alexey Kurakin, Nicolas Papernot, Ian Goodfellow, Dan Boneh,
  and Patrick McDaniel.
\newblock Ensemble adversarial training: Attacks and defenses.
\newblock In {\em Proc. Int'l Conf. on Learning Representations}, Vancouver,
  Canada, May 2018.

\bibitem{Xie}
Cihang Xie, Jianyu Wang, Zhishuai Zhang, Zhou Ren, and Alan Yuille.
\newblock Mitigating adversarial effects through randomization.
\newblock In {\em Proc. Int'l Conf. on Learning Representations}, Vancouver,
  Canada, May 2018.

\bibitem{AdvInputTransform}
Chuan Guo, Mayank Rana, Moustapha Ciss{\'{e}}, and Laurens van~der Maaten.
\newblock Countering adversarial images using input transformations.
\newblock {\em CoRR}, abs/1711.00117, 2017.

\bibitem{Akhtar2018}
Naveed Akhtar, Jian Liu, and Ajmal Mian.
\newblock Defense against universal adversarial perturbations.
\newblock In {\em Proc. {IEEE} Conf. on Computer Vision and Pattern
  Recognition}, pages 3389--3398, Salt Lake City, UT, June 2018.

\end{thebibliography}
% 应该用http://ctan.org/tex-archive/biblio/bibtex/contrib/iopart-num/ 但无法下载，暂时用unsrt.bst

\end{document}